\documentclass[dvipsnames,format=sigconf,anonymous=false,review=false]{article}

\usepackage{arxiv}

\AtBeginDocument{%
  \providecommand\BibTeX{{%
    \normalfont B\kern-0.5em{\scshape i\kern-0.25em b}\kern-0.8em\TeX}}}

\usepackage[utf8]{inputenc} 
\usepackage[T1]{fontenc}    
\usepackage{hyperref}       
\usepackage{url}            
\usepackage{booktabs}       
\usepackage{amsfonts}       
\usepackage{nicefrac}       
\usepackage{microtype}      
\usepackage{lipsum}		
\usepackage{graphicx}
\usepackage{natbib}
\usepackage{doi}

\usepackage{multirow}
\usepackage{amsthm}
\usepackage{comment}

\usepackage{algpseudocode}
\usepackage{algorithm}
\usepackage{makecell}
\usepackage{subcaption}
\usepackage{amsmath}
\usepackage{amssymb}
\usepackage{hyperref}
\usepackage{cleveref}

\hypersetup{draft}

\title{On Scalability of Multi-Objective Evolutionary Algorithms on Combinatorial Optimisation Problems}

\author{
Menghao Tang \\
University of Birmingham \\
Birmingham, United Kingdom \\
\texttt{mxt493@alumni.bham.ac.uk} \\
\And
Zimin Liang \\
University of Birmingham \\
Birmingham, United Kingdom \\
\texttt{zxl525@student.bham.ac.uk} \\
\And
Miqing Li\thanks{Corresponding author} \\
University of Birmingham \\
Birmingham, United Kingdom \\
\texttt{m.li.8@bham.ac.uk} \\
}

\renewcommand{\shorttitle}{On Scalability of MOEAs on MOCOPs}

\begin{document}

\maketitle
\begin{abstract}
  The scalability of evolutionary algorithms refers to how their performance changes as problem size increases. In the multi-objective optimisation field, research on the scalability of multi-objective evolutionary algorithms (MOEAs) has predominantly focussed on continuous problems. 
  However, multi-objective combinatorial optimisation problems (MOCOPs) differ from continuous ones. Their discrete and rigid structure often brings rugged landscape, numerous local optimal solutions, and disjoint global optimal regions. This leads to different behaviour of MOEAs. For example, SEMO, a simple MOEA without mating selection and diversity maintenance mechanisms, has been shown to be highly competitive, and in many cases to outperform more sophisticated MOEAs on MOCOPs. Yet, it remains unclear whether such findings hold for large-scale cases. 
  In this paper, we conduct an empirical investigation into the scalability of MOEAs on combinatorial problems, with problem size from 100 to 5,000. Our results show that SEMO experiences a clear decline in convergence speed as dimensionality increases, compared to other MOEAs such as NSGA-II, SMS-EMOA and MOEA/D. We further demonstrate that the absence of crossover is a major contributor to SEMO's underperformance in large-scale problems, and that incorporating crossover into SEMO can substantially accelerate convergence in general. However, with crossover SEMO may reduce the diversity of the population, failing to spread solutions over the Pareto front.

\end{abstract}


\section{Introduction}
Real-world optimisation problems often require the simultaneous consideration of multiple objectives, and such problems are referred to as multi-objective optimisation problems (MOPs). Depending on the nature of the problem variables, MOPs can be divided into two categories. The first is continuous MOPs, in which decision variables can take any real values within given ranges. The second category is combinatorial MOPs (also referred to as MOCOPs), where decision variables are drawn from a discrete solution space. Due to conflicts among objectives, for both types of MOPs, it is generally impossible to find a single solution that is optimal with respect to all objectives. Instead, the goal is to identify a set of trade-off solutions that are mutually non-dominated, known as Pareto-optimal solutions \cite{Stadler1979}. Multi-objective evolutionary algorithms (MOEAs), due to their population-based nature, are able to approximate the entire set of Pareto-optimal solutions in a single execution, without relying on gradient information or strict mathematical structure. This makes them a natural choice for tackling MOPs.

In many scientific and engineering domains, MOPs involving a large number of decision variables are becoming increasingly common, such as neural network \cite{Castillo2003,Jin2008}, software engineering \cite{Xiang2017,Luna2011,Hierons2016SIP}, scheduling \cite{Jozefowiez2008,Dai2019}, and bioinformatics \cite{Tian2021}. As a result, scalability of MOEAs with respect to the number of decision variables, which describes how algorithmic performance changes as problem size increases, has become an important aspect in measuring MOEAs' effectiveness. To date, several studies have investigated the scalability of evolutionary algorithms on MOPs \cite{Sastry, MiguelAntonio2016, Hong2019}, and these studies consistently reveal a common challenge, ``curse of dimensionality''. For example, empirical studies \cite{Durillo2008, Durillo2010} 
have shown that the efficiency of traditional MOEAs declines sharply as the number of decision variables increases. 

However, most existing scalability studies have been conducted on continuous MOPs. Combinatorial MOPs differ substantially from continuous one. Their discrete and rigid structure often leads to rugged landscapes, numerous local optima, and disjoint global optimal regions. This structural difference can significantly affect algorithmic performance \cite{Li2023}. For instance, a recent comparative study of MOEAs on combinatorial MOPs \cite{Li2024} reported that a simple MOEA without mating selection and diversity preservation, namely SEMO \cite{Laumanns2004}, consistently outperforms several widely used MOEAs (such as NSGA-II \cite{Deb2002}, MOEA/D \cite{Zhang2007} and SMS-EMOA \cite{Beume2007}) on a majority of the considered problems. The experiments in that study were limited to relatively small problem size. Whether SEMO can maintain its advantage as the problem size increases substantially therefore remains an open question.

In addition to the limitation on problem scale, existing studies of MOEAs in combinatorial MOPs tend to be problem-specific. In other words, most study focus on only a single type of combinatorial problems, such as
on the multi-objective knapsack problem (KP) \cite{Ishibuchi2015,Jaszkiewicz2002,Zitzler1999}, travelling salesman problem (TSP) \cite{Shim2011,Psychas2015}, and NK-landscape problem \cite{Merz, Aguirre2007}. In practice, however, different combinatorial MOPs exhibit rather different structures \cite{Liang2025, Fieldsend2025}. For example, the multi-objective NK-landscape problem can be different from the multi-objective TSP and quadratic assignment problem (QAP), resulting in different behaviour of local search heuristics \cite{liang2026random, Behmanesh2020}. In fact, as observed in \cite{Li2024}, MOEAs may behave rather differently across these distinct problem types.

Motivated by the above gaps, this paper conducts an empirical study to investigate the scalability of MOEAs. We choose four algorithms: SEMO \cite{Laumanns2004} which has been demonstrated to perform very competitively in combinatorial MOPs \cite{Li2024}, and three representative well-established MOEAs: NSGA-II \cite{Deb2002}, SMS-EMOA \cite{Beume2007}, and MOEA/D \cite{Zhang2007} on diverse combinatorial MOPs with the problem size scaled from 100 to 5,000. 
One of our most notable findings is that, compared with the three well-established MOEAs, SEMO exhibits a clear decline in convergence speed on large-scale instances. In other words, its search progress becomes increasingly slower than the other three MOEAs as the problem size increases, despite that, providing a sufficient budget, SEMO can typically find a broader Pareto front. A major reason for this behaviour is the absence of crossover in SEMO. 
Incorporating crossover can significantly speed up SEMO's convergence, particularly for large-scale problems. Note that SEMO starts its search with a single randomly generated solution, rather than a population sampled across the search space as in other MOEAs. It is interesting to see that crossover between very similar solutions can still be beneficial.    


The rest of the paper is organised as follows. Chapter 2 introduces the preliminaries of multi-objective optimisation and the MOEAs considered in this study. Chapter 3 describes the experimental design, including the combinatorial MOPs investigated, the quality indicator, and parameter settings. Chapter 4 presents the comparative results across different problem types and sizes. Chapter 5 gives the performance of the SEMO with crossover. Chapter 6 concludes the paper with a summary of findings, and directions for future research.

\section{Multi-Objective Optimisation}

\subsection{Definitions}


Without loss of generality, 
a minimisation MOCOP with $m$ objectives can be formulated as $f(s) = (f_1(s),...,f_m(s))$, where solution $s$ is drawn from a finite decision space $S$ and evaluated by multiple objective functions, resulting in an objective vector in the objective space $Z$ ($Z \subseteq \mathbb{R}^m$).
Due to conflicts between objectives, no single solution can optimise all objectives simultaneously. Therefore, solution quality in multi-objective optimisation is defined based on Pareto dominance. 
Formally, given two objective vectors $z, z'\in Z$, 
$z$ is said to (Pareto) dominate $z'$ (denoted by $z \prec z'$) iff for all $i\in \{1,...,m\}$, $z_i \leq z'_i$, 
and $z \neq z'$.
This dominance relation naturally extends to solutions.
A solution $s$ is said to dominate another solution $s'$ iff $f(s) \prec f(s')$. 
With this definition, a solution $s \in S$ is Pareto optimal if it is not dominated by any other solution in decision space.
The set of Pareto optimal solutions is called the Pareto optimal set, 
and its mapping in the objective space is called the Pareto front.

In practice, for many problems, particularly NP-hard MOCOPs, finding the Pareto front is computationally infeasible \cite{figueira2017easy}.
Therefore, the goal of search algorithms is to find a as good as possible approximation to the Pareto front. 
To evaluate the quality of such approximations, two aspects are typically considered: convergence, which measures the proximity of the obtained solutions to the Pareto front, and diversity (or coverage), which reflects their distribution along the front. An effective multi-objective algorithm is expected to achieve a good balance between convergence and diversity, providing decision-makers with a set of well-distributed solutions.


\subsection{Algorithms Investigated}

In this study, we consider SEMO \cite{Laumanns2004}, which has demonstrated superior performance on MOCOPs in prior work \cite{Li2024}, together with three representative evolutionary algorithms from different paradigms—NSGA-II \cite{Deb2002}, SMS-EMOA \cite{Beume2007} and MOEA/D \cite{Zhang2007}.

\begin{algorithm}[tbp]
\footnotesize
\caption{Simple Evolutionary Multiobjective Optimisation (SEMO) \cite{Laumanns2004}}
\label{Alg:SEMO}
\begin{algorithmic}[1]
\State $s \gets rand\_generation()$
\State $A \gets \{s\}$
\Repeat
    \State $s \gets random\_selection(A)$ \Comment{Select a solution randomly from $A$}
    \State $s' \gets mutation(s)$  \Comment{Generate a new solution using only mutation}
        \If{$A \not\prec s'$}
            \State $A \gets A \cup \{s^\prime\} \setminus \{a \in A \mid s'\prec a\}$
        \EndIf
\Until{$stop\_condition()$}
\Ensure $A$
\end{algorithmic}
\end{algorithm}

\vspace{4pt}
\noindent
\textbf{Simple Evolutionary Multiobjective Optimisation (SEMO)} \cite{Laumanns2004}.
SEMO is a very simple evolutionary algorithms and is frequently used in theoretical research\cite{Bian2018,Doerr2021,Giel2006,Osuna2018,Qian2013,liang2026random}.
As shown in Algorithm~\ref{Alg:SEMO}, it maintains an unbounded archive containing only non-dominated solutions and operates in a ($\mu+1$) style evolutionary manner. In each iteration, SEMO selects one solution at random from the archive and generates a single offspring only through mutation. The offspring is added to the archive if it is not dominated by any existing solution, and any solutions dominated by it are removed. SEMO is strictly elitist, yet it has no explicit population diversity mechanism beyond retaining all non-dominated solutions.

\vspace{4pt}
\noindent
\textbf{Non-dominated Sorting Genetic Algorithm II (NSGA-II)}~\cite{Deb2002}.
NSGA-II is a widely used Pareto-based multi-objective evolutionary algorithm. It employs fast non-dominated sorting to rank solutions and a crowding-distance mechanism to maintain diversity. Elitism is ensured by selecting the best individuals from the combined parent and offspring populations at each generation. Due to its efficiency and robust performance across a wide range of problems, NSGA-II has become one of the most influential baseline algorithms in evolutionary multi-objective optimisation.

\vspace{4pt}
\noindent
\textbf{\emph{S} Metric Selection Evolutionary Multiobjective Optimisation Algorithm (SMS-EMOA)}~\cite{Beume2007}.
SMS-EMOA is an indicator-based evolutionary algorithm that uses hypervolume contribution as the primary selection criterion. At each iteration, it generates a single offspring and removes the individual with the smallest hypervolume contribution from the population, thereby directly optimizing the hypervolume indicator. This indicator-focused selection often provides excellent convergence toward the Pareto front but typically comes with higher computational cost, especially with many objectives \cite{Shang2021,Beume2007}



\vspace{4pt}
\noindent
\textbf{Multiobjective Evolutionary Algorithm Based on Decomposition (MOEA/D)}~\cite{Zhang2007}. 
MOEA/D is a decomposition-based multi-objective evolutionary algorithm that reformulates a multi-objective problem into a set of scalar subproblems using weight vectors. Each subproblem is optimized collaboratively with its neighboring subproblems through solution sharing and localized variation. It represents a conceptually different paradigm compared with Pareto-based and indicator-based methods \cite{Liefooghe2020, Ochoa2023}.

 



It is important to note that the final population of an MOEA does not necessarily reflect its overall search capability \cite{Li2024a,Tanabe2017}. 
In particular, it may contain inferior solutions that are in fact dominated by solutions discarded in earlier population updates \cite{Li2019}. 
Therefore, it is unfair to compare the population of MOEAs with the archive of SEMO.
To enable a fair and comprehensive investigation of the search performance, we use an external archive to record all non-dominated solutions generated during the search for NSGA-II, SMS-EMOA and MOEA/D, and compare their archives with SEMO's. 

\section{Experimental Design}

\subsection{Optimisation Problems Considered}
We consider the four well-established multi-objective combinatorial optimisation problems used in \cite{Li2024}.

\vspace{4pt}
\noindent
\textbf{Multi-Objective Travelling Salesman Problem (MOTSP)}.
The multi-objective traveling salesman problem \cite{Borges2002} generalizes the classical traveling salesman problem by incorporating multiple cost criteria. Let $V={{v}_1,\ldots,v_D}$ denote a set of $D$ cities. A feasible solution $x$ to the TSP is a Hamiltonian cycle, i.e., a permutation of all cities that forms a closed tour visiting each city exactly once. Instead of a single distance or cost function, the MOTSP considers $m$ objective functions, each defined by a cost matrix $C_j=\left(C_{j,uv}\right)_{D\times D}$, where $C_{j,uv}$ represents the cost of traveling from city $u$ to city $v$ under objective $j$. The objective of the MOTSP is to minimize simultaneously the $m$ total travel costs associated with the tour. Like in \cite{Corne2007}, the $m$ cost matrices are independently generated by assigning each pair of cities with a number randomly drawn from [0, 1).

\vspace{4pt}
\noindent
\textbf{Multi-Objective 0-1 Knapsack Problem (MOKP)} 
The multi-objective 0-1 knapsack problem \cite{Zitzler1999} is a widely used MOCOP. The problem is defined over a set of $D$ items and $m$ knapsack. Each item $i$ has an associated value $v_{ji}$ and weight $w_{ji}$ for each knapsack $j=1,\ldots,m$. Selecting an item means that it is placed into all $m$ knapsacks simultaneously. Each knapsack $j$ has its own capacity limit $c_j$. The objective is to select a subset of items that maximizes the $m$ value objectives while ensuring that none of the $m$ knapsack capacities is exceeded. Formally, this problem can be defined as follows:
\begin{equation}
\begin{split}
\textrm{max}~f_{j}{(x)} = \sum_{i=1}^{D}v_{ji}x_{i},~~j=1,...,m \\
\textrm{subject to}~ \sum_{i=1}^{D}w_{ji}x_{i} \leq c_{j}, ~~j=1,...,m \\
\end{split}
\label{eq:Knapsack}
\end{equation}
A solution is represented by a binary vector $x=\left(x_1,\ldots,x_D\right)$, where $x_i\in\{0,1\}$, $x_i=1$ means that item $i$ is selected. As in \cite{Zitzler1999}, each problem instance is generated as follows: $v_{ji}$ and $w_{ji}$ are drawn uniformly at random from the integer interval $[10, 100]$, and the knapsack capacity is set to half of the total weight of all items. This problem is also known as the multi-objective multidimensional knapsack problem, or MOMKP.

\vspace{4pt}
\noindent
\textbf{Multi-Objective NK-Landscape Problem (MONK)}
The NK- Landscape problem \cite{Aguirre2004} is widely used in multi-objective optimisation because its landscape ruggedness can be controlled through the parameter $K$. In this problem, a candidate solution is represented by a binary string of length $D$, $x=\left(x_1,\ldots,x_D\right)\in\{0,1\}^D$. Each bit in the string contributes to the objective values, and this contribution depends not only on the bit itself but also on $K$ other interacting bits. For each objective $f_j$, every bit $x_i$ is associated with a contribution function $c_{i,j}\left(x_i,x_{k_{i,j,1}},\ldots,x_{k_{i,j,K}}\right)$, where index $k_{i,j,1}$ denotes the first interacting position that influences the contribution of bit $i$ under objective $j$. The problem can be defined as follows:
\begin{equation}
\textrm{max}~ f_j(x)
= \frac{1}{D} \sum_{i=1}^{D}
c_{i,j}(x_i, x_{k_{i,j,1}}, \ldots, x_{k_{i,j,K}}),~~
j = 1,\ldots,m
\label{eq:MONK}
\end{equation}
In our experiments, $K$ was set  to 10.
Following \cite{Aguirre2007,Daolio2015}, the fitness contribution $c_{i,j}$ is randomly drawn from $[0, 1)$ for every possible combination $\{0,1\}^{\left(K+1\right)}$ of the interacting bits $k_{i,j}$.

\vspace{4pt}
\noindent

\textbf{Multi-Objective Quadratic Assignment Problem (MOQAP)}.
The multi-objective quadratic assignment problem \cite{Knowles2003} involves assigning $D$ facilities to $D$ locations, where the cost of an assignment depends on the pairwise interactions between facilities and the distances between their assigned locations. For each objective $j=1,\ldots,m$, a flow matrix ${\ C}_j=\left(C_{j,uv}\right)_{D\times D}$ specifies the interaction cost between facilities $u$ and $v$, and a distance matrix $L=\left(L_{pq}\right)_{D\times D}$ defines the distance between locations $p$ and $q$. A solution is represented by a permutation $\pi$ of ${1,\ldots,D}$, where $\pi\left(u\right)$ denotes the location assigned to facility $u$. The aim is to identify assignments that minimize all $m$ objective functions simultaneously. Formally, for a given permutation $\pi$, the objective functions are defined as:
\begin{equation}
\textrm{min}~ f_j(\pi)
= \sum_{u=1}^{D} \sum_{v=1}^{D}
C_{j,uv}\, L_{\pi(u),\,\pi(v)},~~ j = 1,\ldots,m
\label{eq:MOQAP}
\end{equation}
Following \cite{Knowles2003}, the $m$ cost matrices are independently generated by sampling each entry from $\left[0,100\right]$. And the distance matrix is constructed, as in \cite{Taillard1995}, by randomly placing $D$ locations in a $\left[0,5000\right]^2$ plane and computing pairwise Euclidean distances.

\subsection{Quality Indicator Used}

To assess the solution quality obtained by each algorithm, we employ the hypervolume (HV) \cite{Zitzler1999} as quality indicator. The HV measures the volume of the objective space dominated by the non-dominated set and bounded by a reference point. It can reflect convergence, spread, uniformity and cardinality of a solution set, making it particularly suitable for comparing MOEAs when the problem's Pareto front is unknown \cite{Li2019a,Shang2021}.

In practice, the HV values obtained by different algorithms can be highly sensitive to the choice of the reference point. In particular, if a solution lies outside the region defined by the reference point, it contributes zero to the HV value. A common approach is to define the reference point slightly worse (e.g., $\times 1.1$) than the nadir point of the Pareto front, or, when the Pareto front is unavailable, to approximate it using the nadir point of the nondominated solutions set generated by all algorithms under all the settings.

However, these approaches still cannot avoid this issue when there is a large performance gap among algorithms, which is the case in combinatorial MOPs. In such cases, poorly performing algorithms may fall far outside the reference region, resulting in situation where only the best-performing algorithm attains a non-zero HV value. This issue indeed occurs in our study, particularly for large-scale problem instances.
To address it, we follow the approach in \cite{Li2024}.
We use random sampling to determine the reference point. For each problem, all nondominated solutions generated by the random sampling procedure are collected, and the nadir point of this set is identified for each objective dimension. The reference point is then defined to be slightly worse than the nadir point. According to \cite{Li2022}, for maximisation problems, the reference point is set as $r_i = min_i - (max_i-min_i)/10$, and similarly for minimisation problems as $r_i = max_i + (max_i-min_i)/10$, where $max_i$ and $min_i$ denote the maximum and minimum values of the nondominated set on the $i$th objective, respectively.

\subsection{General Experimental Settings}

In this experiment, all algorithms are tested under the same experimental framework, using consistent operators and parameter settings to each problem type, to ensure a fair comparison. To investigate scalability, the experiments cover a wide range of problem scales, with the number of decision variables $D$ set to 100, 500, 1000, and 5000, unless stated otherwise. Each algorithm is evaluated under two search budgets, $10^5$ and $10^7$ function evaluations, to assess performance under a common budget and to examine how performance changes when the budget is increased substantially.


The search operators are selected in accordance with the encoding structure of each problem. For binary-encoded problems such as MOKP and MONK, NSGA-II, SMS-EMOA, and MOEA/D employ uniform crossover \cite{Eiben2015} with a probability of $p_c=1.0$ and bit-flip mutation \cite{Eiben2015} with $p_m = 1/D$, where $D$ denotes the number of variables. In contrast, SEMO only employs local search operators. Following the practice in \cite{Li2024}, we set a one-bit flip for MONK, while a two-bit flip is adopted for MOKP. The reason we consider two-bit flip for MOKP is that flipping a single bit usually leads to either improvement or deterioration across all objectives, hence not suitable local moves. 
For MOTSP, which is based on order-related permutations, NSGA-II, SMS-EMOA, and MOEA/D employ order-based crossover combined with 2-opt mutation \cite{Eiben2015}. For MOQAP, which is based on position-related permutations, cycle crossover together with a 2-swap mutation \cite{Eiben2015} is adopted. SEMO uses the same mutation operators, 2-opt for MOTSP and 2-swap for MOQAP. For both permutation problems, the crossover rate is set to 1.0 and the mutation rate to 0.05.

Unless otherwise stated, all remaining algorithm parameters follow the recommendations in the original papers. For NSGA-II, SMS-EMOA, and MOEA/D, the population size is set to 100. 
For each problem instance, each algorithm is executed independently for 30 runs to ensure statistically reliable results.

\section{Results}
In this section, we report the experimental results and their statistical analysis based on hypervolume (HV) values. To rigorously assess overall performance differences among the algorithms, the Friedman rank-sum test \cite{Friedman1940} was applied at a significance level of 0.05, as it is well suited for non-parametric comparisons involving multiple groups. The results indicate statistically significant differences among the algorithms in all cases. Given this outcome, we further conducted pairwise analyses to examine the relative performance of individual algorithms using the Wilcoxon rank-sum test \cite{Haynes2013} at the same significance level. To avoid the family-wise error arising from multiple comparisons, Holm method \cite{Holm1979} was employed to adjust the p-values obtained from the Wilcoxon tests.

\Cref{tab:MOKP,tab:MOTSP,tab:MOQAP,tab:MONK} present, for each of the four MOCOPs, the mean and standard deviation of the HV values obtained from 30 independent runs for each algorithm, together with the numbers of pairwise comparisons in which the algorithm performs statistically better, equal, or worse than the other algorithms.

\subsection{On the Travelling Salesman Problem}

\begin{table*}[!htbp]
\centering
\caption{HV results (mean and SD) of the four algorithms on the multi-objective TSP (MOTSP) under the eight settings. The bold numbers for each algorithm denote the numbers of pairwise comparisons in which the algorithm performs statistically better, equally, or worse than its competitors, respectively.}
\vspace{-5pt}
\footnotesize
\begin{tabular}{clcccc}
\hline
Dimension &
  \multicolumn{1}{c}{Budget} &
  SEMO &
  NSGA-II &
  SMS-EMOA &
  MOEA/D \\ \hline

\multicolumn{1}{c|}{\multirow{4}{*}{100D}} &
  \multicolumn{1}{l|}{\multirow{2}{*}{$10^5$ evals}} &
  2.18e+03 (1.88e+01)&
  1.83e+03 (6.24e+01) &
  1.87e+03 (6.49e+01) &
  1.95e+03 (4.73e+01)\\
\multicolumn{1}{c|}{} &
  \multicolumn{1}{l|}{} &
  \textbf{3/0/0} &
  \textbf{0/0/3} &
  \textbf{1/0/2} &
  \textbf{2/0/1} \\ \cline{2-6}

\multicolumn{1}{c|}{} &
  \multicolumn{1}{l|}{\multirow{2}{*}{$10^7$ evals}} &
  2.69e+03 (5.65e+00) &
  2.57e+03 (1.30e+01) &
  2.53e+03 (2.16e+01) &
  2.52e+03 (1.91e+01) \\
\multicolumn{1}{c|}{} &
  \multicolumn{1}{l|}{} &
  \textbf{3/0/0} &
  \textbf{2/0/1} &
  \textbf{0/1/2} &
  \textbf{0/1/2} \\ \hline

\multicolumn{1}{c|}{\multirow{4}{*}{500D}} &
  \multicolumn{1}{l|}{\multirow{2}{*}{$10^5$ evals}} &
  2.63e+04 (6.31e+02) &
  2.16e+04 (1.32e+03) &
  2.45e+04 (1.23e+03) &
  2.75e+04 (6.40e+02) \\
\multicolumn{1}{c|}{} &
  \multicolumn{1}{l|}{} &
  \textbf{2/0/1} &
  \textbf{0/0/3} &
  \textbf{1/0/2} &
  \textbf{3/0/0} \\ \cline{2-6}

\multicolumn{1}{c|}{} &
  \multicolumn{1}{l|}{\multirow{2}{*}{$10^7$ evals}} &
  5.54e+04 (1.45e+02) &
  5.03e+04 (8.83e+02) &
  5.06e+04 (6.12e+02) &
  5.29e+04 (3.86e+02) \\
\multicolumn{1}{c|}{} &
  \multicolumn{1}{l|}{} &
  \textbf{3/0/0} &
  \textbf{0/1/2} &
  \textbf{0/1/2} &
  \textbf{2/0/1} \\ \hline

\multicolumn{1}{c|}{\multirow{4}{*}{1000D}} &
  \multicolumn{1}{l|}{\multirow{2}{*}{$10^5$ evals}} &
  6.61e+04 (2.57e+03) &
  6.09e+04 (3.93e+03) &
  6.92e+04 (3.80e+03) &
  8.05e+04 (2.22e+03) \\
\multicolumn{1}{c|}{} &
  \multicolumn{1}{l|}{} &
  \textbf{1/0/2} &
  \textbf{0/0/3} &
  \textbf{2/0/1} &
  \textbf{3/0/0} \\ \cline{2-6}

\multicolumn{1}{c|}{} &
  \multicolumn{1}{l|}{\multirow{2}{*}{$10^7$ evals}} &
  1.90e+05 (6.44e+02) &
  1.73e+05 (4.69e+03) &
  1.85e+05 (4.68e+03) &
  1.97e+05 (1.24e+03) \\
\multicolumn{1}{c|}{} &
  \multicolumn{1}{l|}{} &
  \textbf{2/0/1} &
  \textbf{0/0/3} &
  \textbf{1/0/2} &
  \textbf{3/0/0} \\ \hline

\multicolumn{1}{c|}{\multirow{4}{*}{5000D}} &
  \multicolumn{1}{l|}{\multirow{2}{*}{$10^5$ evals}} &
  1.82e+05 (1.81e+04) &
  4.31e+05 (3.02e+04) &
  6.43e+05 (2.92e+04) &
  6.85e+05 (1.54e+04) \\
\multicolumn{1}{c|}{} &
  \multicolumn{1}{l|}{} &
  \textbf{0/0/3} &
  \textbf{1/0/2} &
  \textbf{2/0/1} &
  \textbf{3/0/0} \\ \cline{2-6}

\multicolumn{1}{c|}{} &
  \multicolumn{1}{l|}{\multirow{2}{*}{$10^7$ evals}} &
  1.33e+06 (1.17e+04) &
  2.71e+06 (1.35e+05)&
  3.06e+06 (1.45e+05) &
  3.94e+06 (2.16e+04) \\
\multicolumn{1}{c|}{} &
  \multicolumn{1}{l|}{} &
  \textbf{0/0/3} &
  \textbf{1/0/2} &
  \textbf{2/0/1} &
  \textbf{3/0/0} \\ \hline

\end{tabular}
\label{tab:MOTSP}%
\end{table*}

\begin{figure}[tbp]
\vspace{-0pt}
	\begin{center}
        \includegraphics[scale=0.35, trim=0 15 0 15]{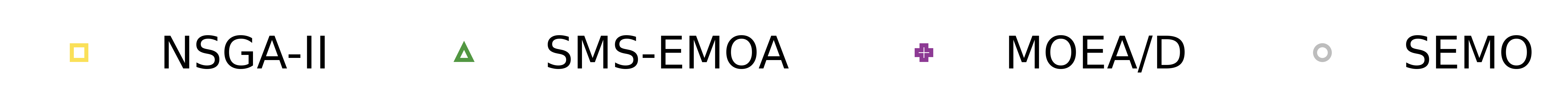}
        \vspace{-0pt}
        \begin{tabular}{@{}cc@{}}
    \includegraphics[scale=0.31]{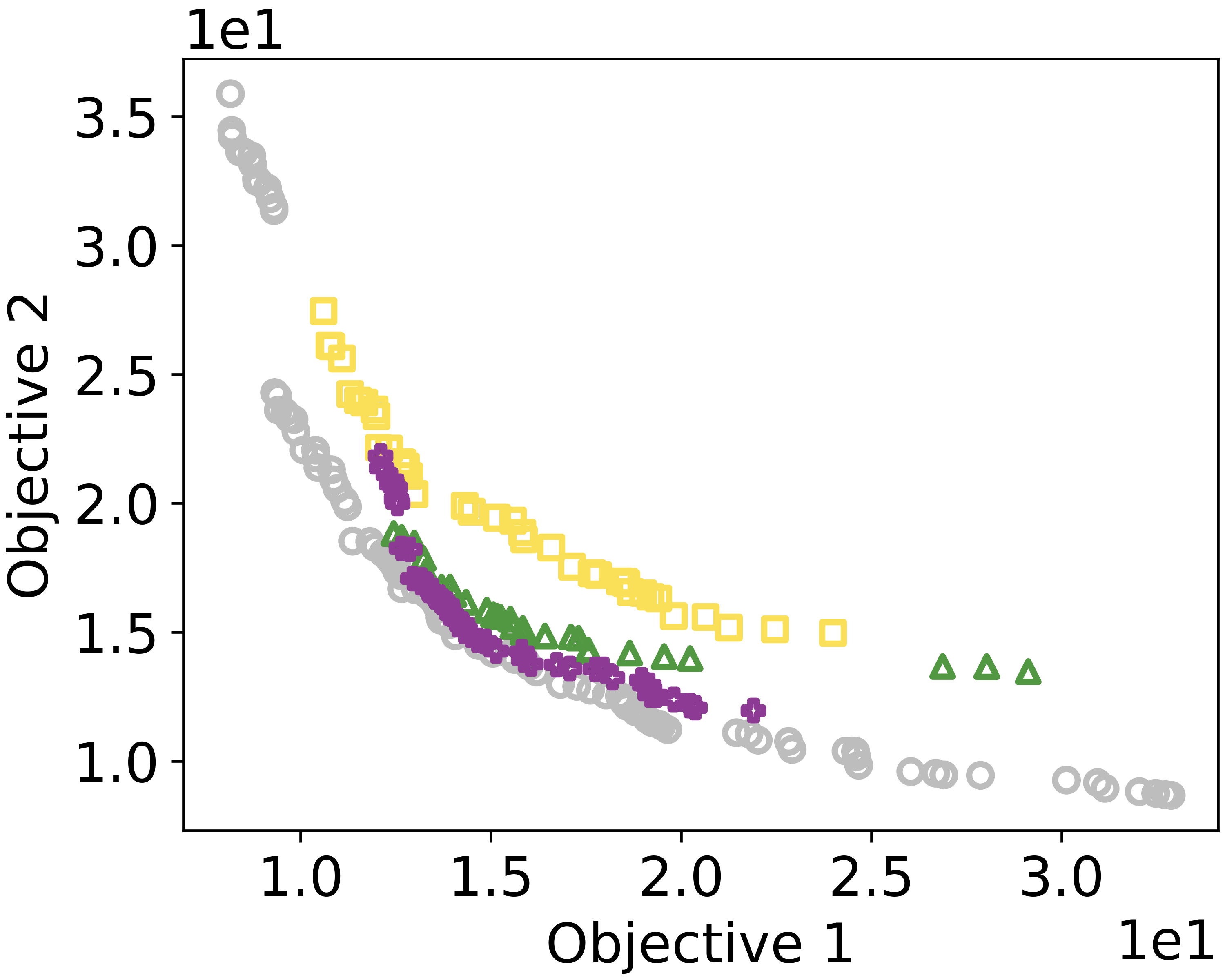} &
    \includegraphics[scale=0.32]{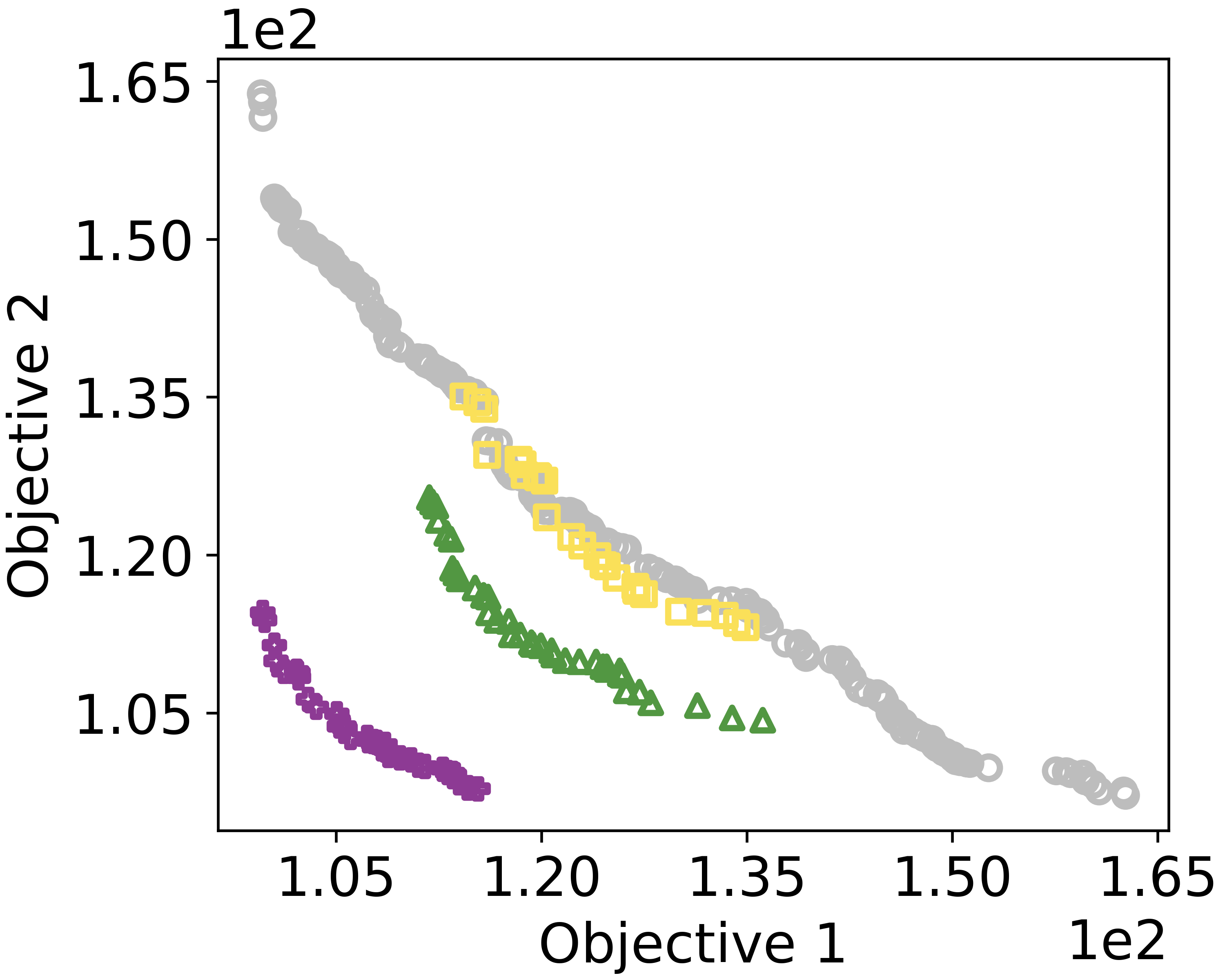} \\
    (a) 100D & (b) 500D \\

    \includegraphics[scale=0.31]{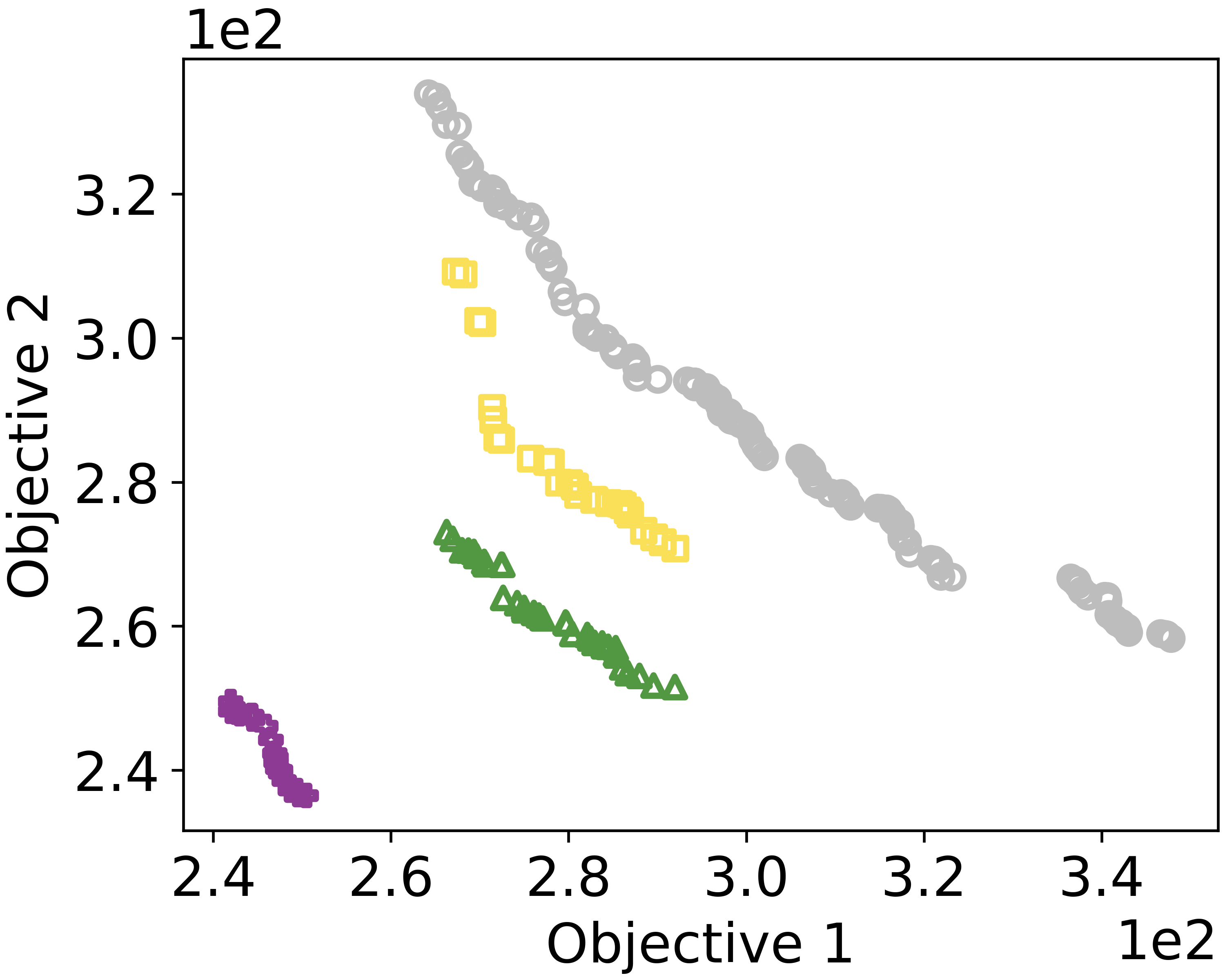} &
    \includegraphics[scale=0.31]{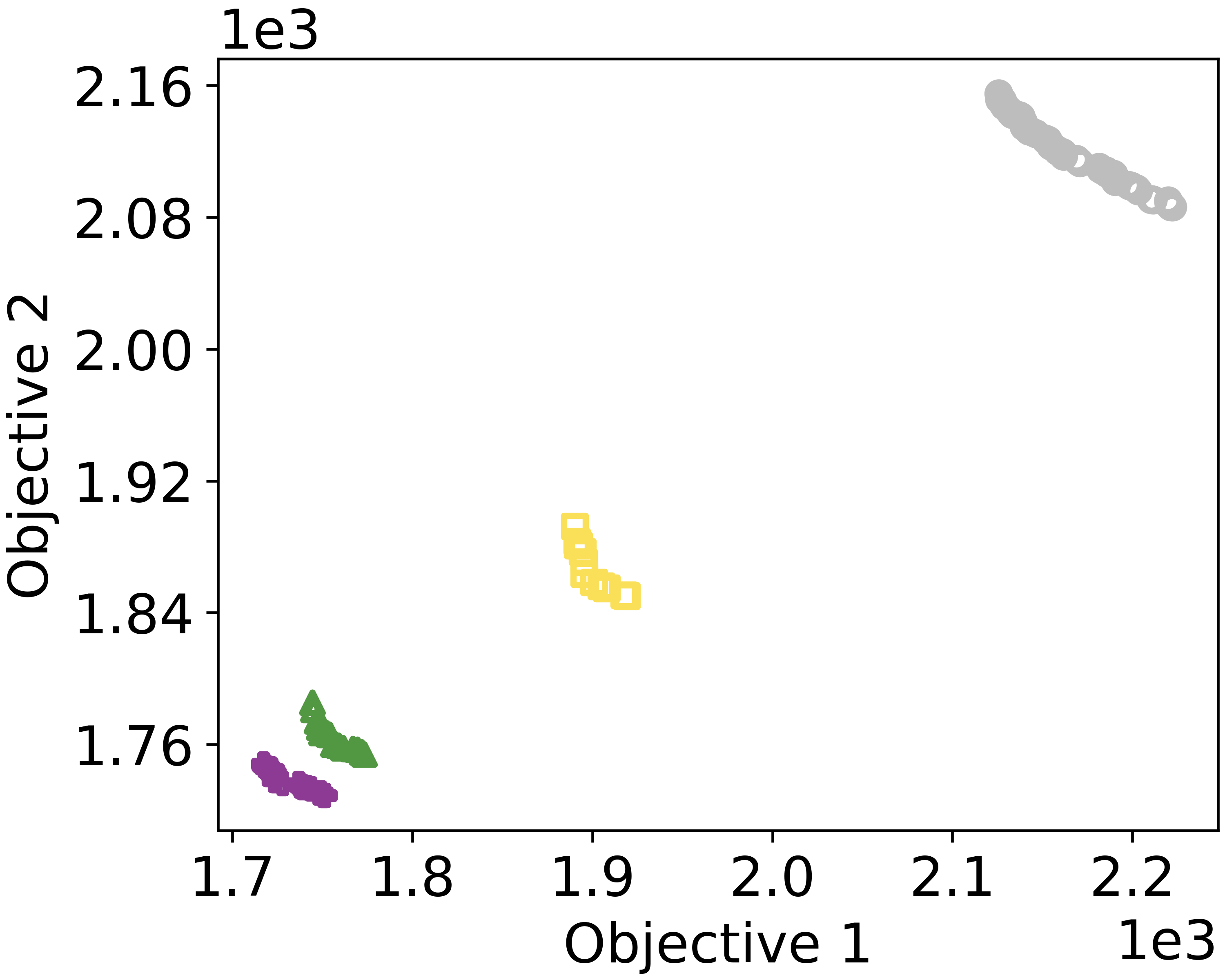} \\
    (c) 1000D & (d) 5000D
    \end{tabular}
	\end{center}
        \vspace{-5pt}
	\caption{Solution sets obtained by the four algorithms in a representative run under a budget of $10^5$ evaluations on MOTSP with (a) 100, (b) 500, (c) 1,000 and (d) 5,000 decision variables.}
	\label{Fig:MOTSP-1}
\end{figure}

\begin{figure}[tbp]
\vspace{-0pt}
	\begin{center}
        \includegraphics[scale=0.35, trim=0 15 0 15]{Figures/Scalability/Legned.png}
        \vspace{-0pt}
        \begin{tabular}{@{}cc@{}}
			\includegraphics[scale=0.31]{Figures/Scalability/TSP-1000-1.png} &
            \includegraphics[scale=0.31]{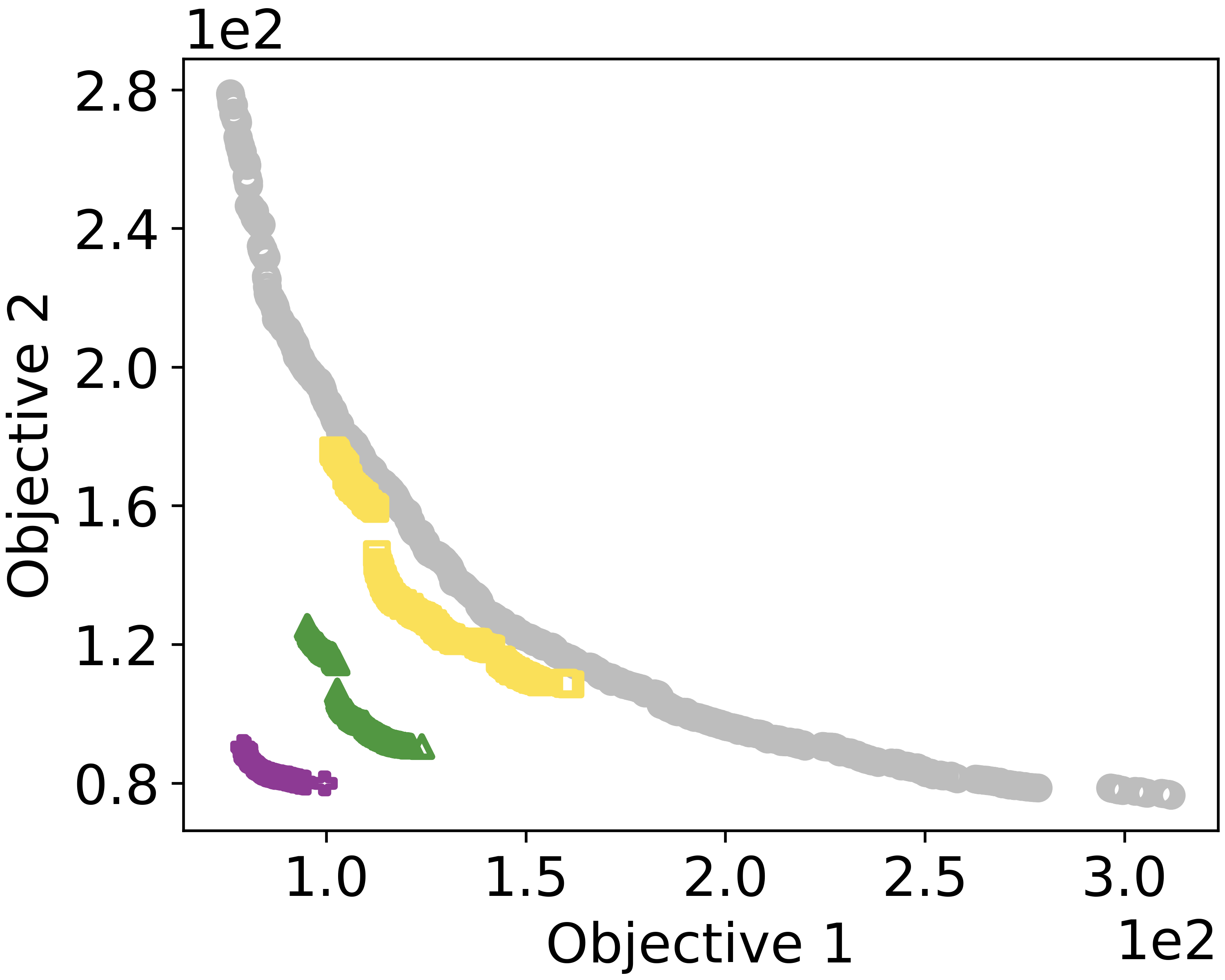}
             \\
			    (a) $10^5$ evaluations&
                (b) $10^7$ evaluations
		\end{tabular}
	\end{center}
        \vspace{-5pt}
	\caption{Solution sets obtained by the four algorithms in a representative run under two budgets of $10^5$ and $10^7$ evaluations on MOTSP with 1,000 decision variables. 
    }
	\label{Fig:MOTSP-2}
\end{figure}
Let us first consider the small budget case (i.e., $10^5$ evaluations). As can be seen from Table~\ref{tab:MOTSP}, as the problem size increases, the performance of SEMO deteriorates progressively compared with the other algorithms. When $D = 100$, SEMO is the best-performing algorithm. However, at $D = 500$, SEMO is outperformed by MOEA/D, and at $D = 1000$, it is overtaken by SMS-EMOA and MOEA/D. When $D = 5000$, SEMO becomes the worst-performing algorithm. 
Figure~\ref{Fig:MOTSP-1} shows the solution sets obtained by the four algorithms in a run on the four problem size. This particular run, along with others for visual demonstration in the paper, is associated with the result which is the closest to the mean HV value. 
As shown, as the problem size increases, SEMO starts to lag behind the other algorithms in convergence, and the convergence gap continues to widen. 

For the large budget case (i.e., $10^7$ evaluations), a similar pattern can be seen from the table. SEMO performs best on the 100D and 500D instances, but is overtaken by MOEA/D on the 1000D instance and becomes the worst algorithm on the 5000D instance. 
Comparing the results obtained under the small and large budgets, one can see that SEMO tends to benefit from larger budgets. That is, although it converges more slowly, SEMO can catch up and achieve a broader Pareto front when given sufficient budget. Figure~\ref{Fig:MOTSP-2} shows such a case, where the solution sets obtained by the four algorithms on the 1000D instance under the two budgets are plotted. Clearly, SEMO falls behind under the $10^5$ evaluations. However, when the budget increases to $10^7$ evaluations, the gap becomes smaller, and SEMO almost catches up with NSGA-II. That said, for the problem size $D = 5000$, much more evaluations are needed for SEMO to catch up as can be seen in Table~\ref{tab:MOTSP}, where SEMO' hypervolume is significantly lower than those of the other three algorithms.

\subsection{On the Knapsack Problem}
\begin{figure}[tbp]
\vspace{-5pt}
	\begin{center}
		\hspace*{-0pt}
        \includegraphics[scale=0.35, trim=0 15 0 15]{Figures/Scalability/Legned.png}
        \vspace{-5pt}
        \begin{tabular}{@{}cc@{}}
			\includegraphics[scale=0.31]{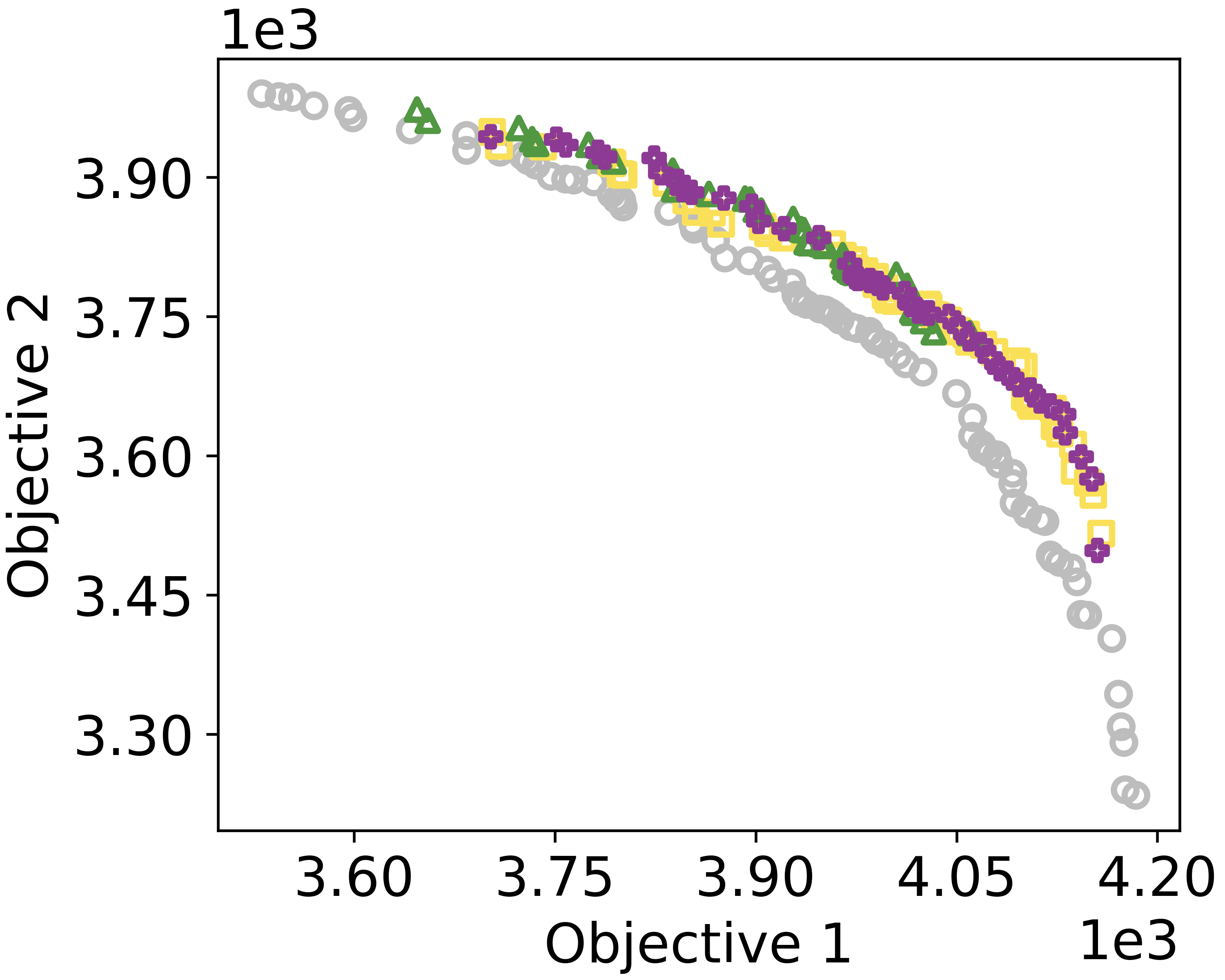} &
            \includegraphics[scale=0.31]{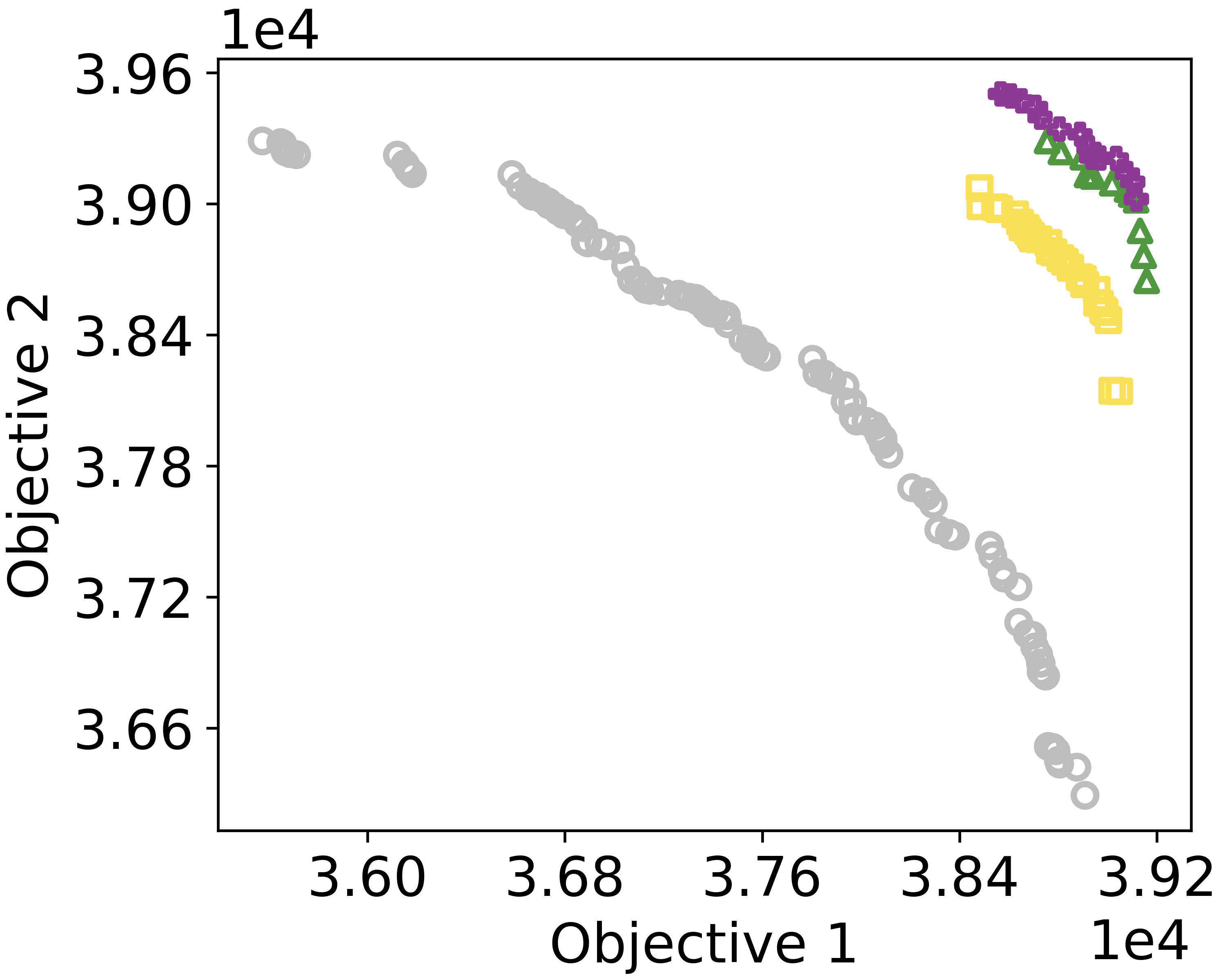}
             \\
			    (a) 100D &
                (b) 1000D
		\end{tabular}
	\end{center}
        \vspace{-5pt}
	\caption{Solution sets obtained by the four algorithms in a representative run under a budget of $10^5$ evaluations on MOKP with (a) 100 and (b) 1,000 decision variables.}
	\label{Fig:MOKP-1}
\end{figure}

\begin{figure}[tbp]
\vspace{-5pt}
	\begin{center}
        \includegraphics[scale=0.35, trim=0 15 0 15]{Figures/Scalability/Legned.png}
        \vspace{-0pt}
        \begin{tabular}{@{}cc@{}}
			\includegraphics[scale=0.31]{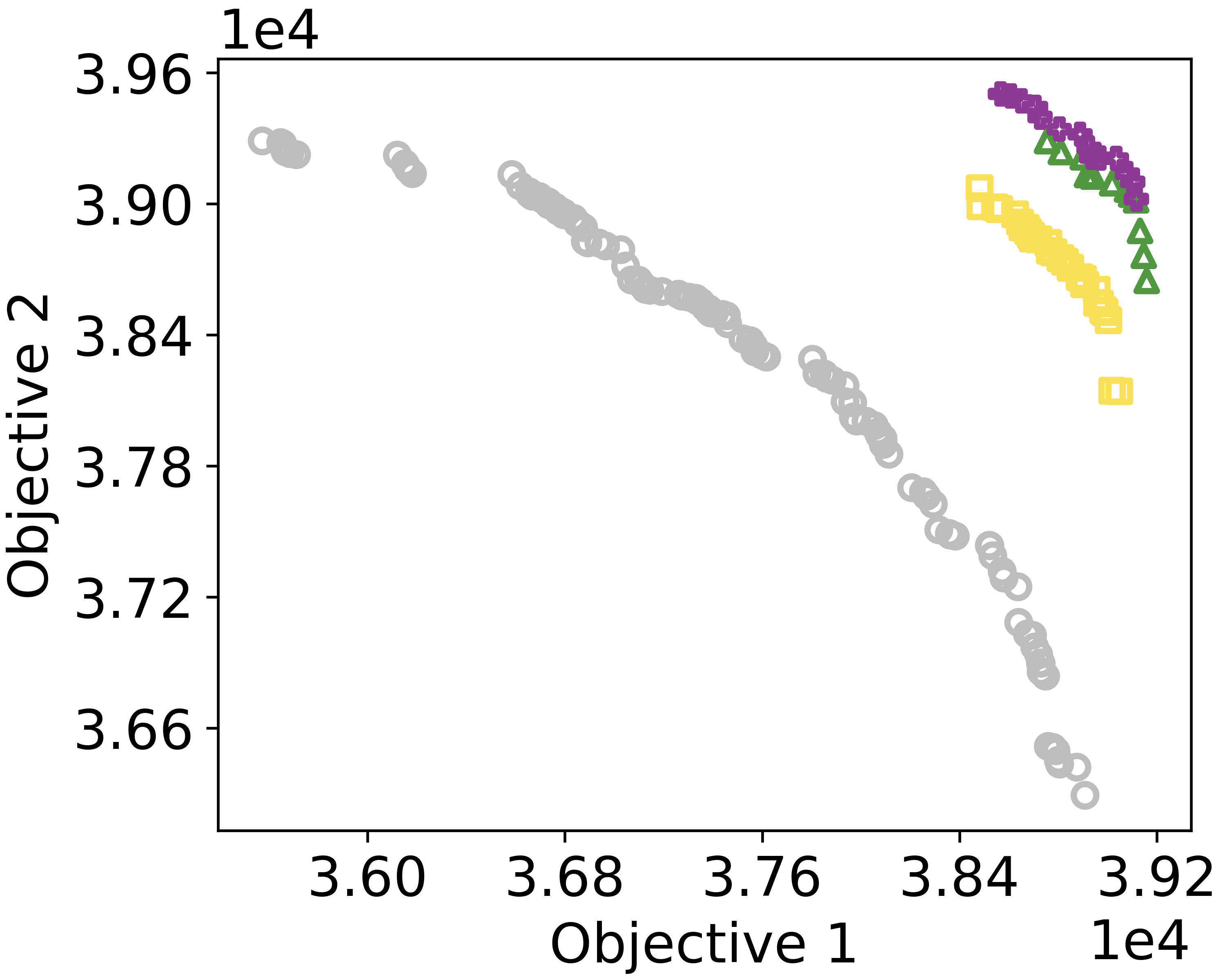} &
            \includegraphics[scale=0.31]{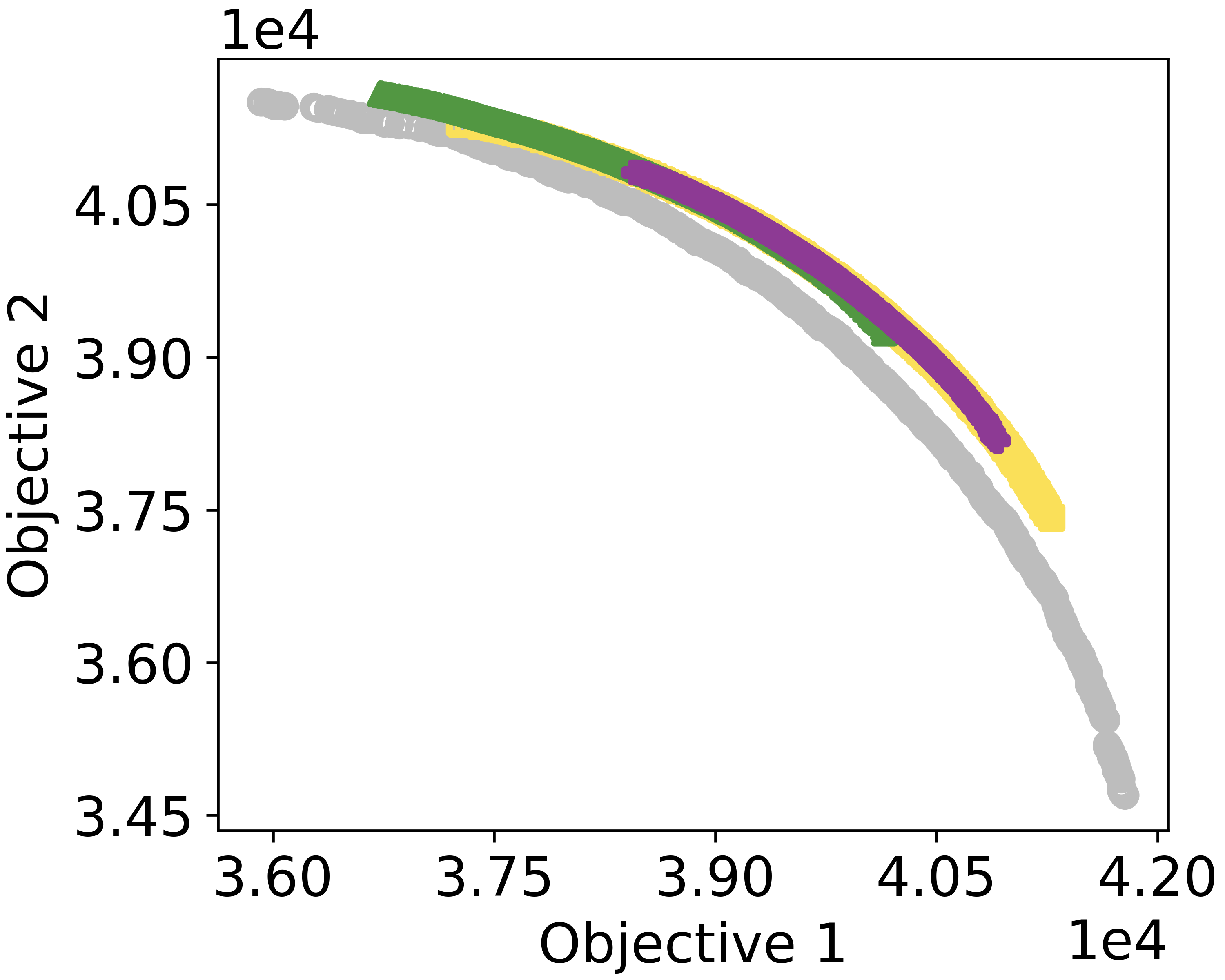}
             \\
			    (a) $10^5$ evaluations&
                (b) $10^7$ evaluations
		\end{tabular}
	\end{center}
        \vspace{-5pt}
	\caption{Solution sets obtained by the four algorithms in a representative run under two budgets of $10^5$ and $10^7$ evaluations on MOKP with 1,000 decision variables. }
	\label{Fig:MOKP-2}
\end{figure}

On the multi-objective knapsack problem (MOKP), SEMO shows a similar performance pattern to that on the MOTSP. Figure~\ref{Fig:MOKP-1} shows the solution sets obtained by the four algorithms on the 100D and 1000D instances. As shown in Table~\ref{tab:MOKP} and Figure~\ref{Fig:MOKP-1}, under both budget settings, SEMO increasingly lags behind the other algorithms as the problem size grows, resulting in a widening gap in convergence. And again, as more evaluations are available, the performance of SEMO improves more rapidly. Figure~\ref{Fig:MOKP-2} illustrates this effect: on the 1000D instance, the solutions obtained by SEMO catch up when the evaluation budget increases to $10^7$. 

\begin{table*}[htbp]
\centering
\caption{HV results (mean and SD) of the four algorithms on the multi-objective knapsack problem (MOKP) under the eight settings. The bold numbers for each algorithm denote the numbers of pairwise comparisons in which the algorithm performs statistically better, equally, or worse than its competitors, respectively.}
\vspace{-5pt}
\footnotesize
\begin{tabular}{clcccc}
\hline
Dimension &
  \multicolumn{1}{c}{Budget} &
  SEMO &
  NSGA-II &
  SMS-EMOA &
  MOEA/D \\ \hline

\multicolumn{1}{c|}{\multirow{4}{*}{100D}} &
  \multicolumn{1}{l|}{\multirow{2}{*}{$10^5$ evals}} &
  2.51e+06 (3.29e+04) &
  2.45e+06 (3.74e+04) &
  2.38e+06 (5.20e+04) &
  2.46e+06 (2.92e+04) \\
\multicolumn{1}{c|}{} &
  \multicolumn{1}{l|}{} &
  \textbf{3/0/0} &
  \textbf{1/1/1} &
  \textbf{0/0/3} &
  \textbf{1/1/1} \\ \cline{2-6}

\multicolumn{1}{c|}{} &
  \multicolumn{1}{l|}{\multirow{2}{*}{$10^7$ evals}} &
  2.57e+06 (2.18e+04) &
  2.60e+06 (7.03e+03) &
  2.60e+06 (8.05e+03) &
  2.59e+06 (5.88e+03) \\
\multicolumn{1}{c|}{} &
  \multicolumn{1}{l|}{} &
  \textbf{0/0/3} &
  \textbf{2/1/0} &
  \textbf{2/1/0} &
  \textbf{1/0/2} \\ \hline

\multicolumn{1}{c|}{\multirow{4}{*}{500D}} &
  \multicolumn{1}{l|}{\multirow{2}{*}{$10^5$ evals}} &
  4.13e+07 (6.61e+05) &
  4.07e+07 (6.78e+05) &
  4.05e+07 (5.94e+05) &
  4.16e+07 (4.66e+05) \\
\multicolumn{1}{c|}{} &
  \multicolumn{1}{l|}{} &
  \textbf{2/1/0} &
  \textbf{0/1/2} &
  \textbf{0/1/2} &
  \textbf{2/1/0} \\ \cline{2-6}

\multicolumn{1}{c|}{} &
  \multicolumn{1}{l|}{\multirow{2}{*}{$10^7$ evals}} &
  5.01e+07 (1.85e+05) &
  4.99e+07 (1.82e+05) &
  4.77e+07 (4.80e+05) &
  4.84e+07 (2.23e+05) \\
\multicolumn{1}{c|}{} &
  \multicolumn{1}{l|}{} &
  \textbf{3/0/0} &
  \textbf{2/0/1} &
  \textbf{0/0/3} &
  \textbf{1/0/2} \\ \hline

\multicolumn{1}{c|}{\multirow{4}{*}{1000D}} &
  \multicolumn{1}{l|}{\multirow{2}{*}{$10^5$ evals}} &
  1.50e+08 (2.92e+06) &
  1.49e+08 (3.14e+06) &
  1.56e+08 (2.09e+06) &
  1.58e+08 (2.37e+06) \\
\multicolumn{1}{c|}{} &
  \multicolumn{1}{l|}{} &
  \textbf{0/1/2} &
  \textbf{0/1/2} &
  \textbf{2/0/1} &
  \textbf{3/0/0} \\ \cline{2-6}

\multicolumn{1}{c|}{} &
  \multicolumn{1}{l|}{\multirow{2}{*}{$10^7$ evals}} &
  2.11e+08 (6.51e+05) &
  2.07e+08 (1.24e+06) &
  1.97e+08 (1.78e+06) &
  1.97e+08 (1.40e+06) \\
\multicolumn{1}{c|}{} &
  \multicolumn{1}{l|}{} &
  \textbf{3/0/0} &
  \textbf{2/0/1} &
  \textbf{0/1/2} &
  \textbf{0/1/2} \\ \hline

\multicolumn{1}{c|}{\multirow{4}{*}{5000D}} &
  \multicolumn{1}{l|}{\multirow{2}{*}{$10^5$ evals}} &
  1.15e+09 (7.88e+07) &
  1.09e+09 (4.57e+07) &
  1.66e+09 (5.50e+07) &
  1.60e+09 (4.68e+07) \\
\multicolumn{1}{c|}{} &
  \multicolumn{1}{l|}{} &
  \textbf{1/0/2} &
  \textbf{0/0/3} &
  \textbf{3/0/0} &
  \textbf{2/0/1} \\ \cline{2-6}

\multicolumn{1}{c|}{} &
  \multicolumn{1}{l|}{\multirow{2}{*}{$10^7$ evals}} &
  3.03e+09 (1.39e+07) &
  4.25e+09 (2.05e+07) &
  4.16e+09 (2.51e+07) &
  4.09e+09 (1.54e+07) \\
\multicolumn{1}{c|}{} &
  \multicolumn{1}{l|}{} &
  \textbf{0/0/3} &
  \textbf{3/0/0} &
  \textbf{2/0/1} &
  \textbf{1/0/2} \\ \hline

\end{tabular}
\label{tab:MOKP}%
\end{table*}

An interesting exception is observed for the instance with 100 decision variables. Under the small budget, SEMO achieves the best performance among all algorithms, whereas under the large budget it is overtaken by the other methods and becomes the worst-performing algorithm. A possible explanation is that, compared with the MOTSP, 100 decision variable represents a relatively small problem scale for the MOKP. With an evaluation budget of $10^7$, all the algorithms are able to obtain solution sets that are very close the Pareto front. In this setting, SEMO which relies solely on a local search operator may become trapped into local optima and thus fail to make further progress. In contrast, the other three MOEAs can continue to advance through their global variation operators of crossover and mutation when sufficient evaluations are available. 


\subsection{On the NK-landscape Problem}
\begin{figure}[tbp]
\vspace{-5pt}
	\begin{center}
		\hspace*{-0pt}
        \includegraphics[scale=0.35, trim=0 15 0 15]{Figures/Scalability/Legned.png}
        \vspace{-5pt}
        \begin{tabular}{@{}cc@{}}
			\includegraphics[scale=0.31]{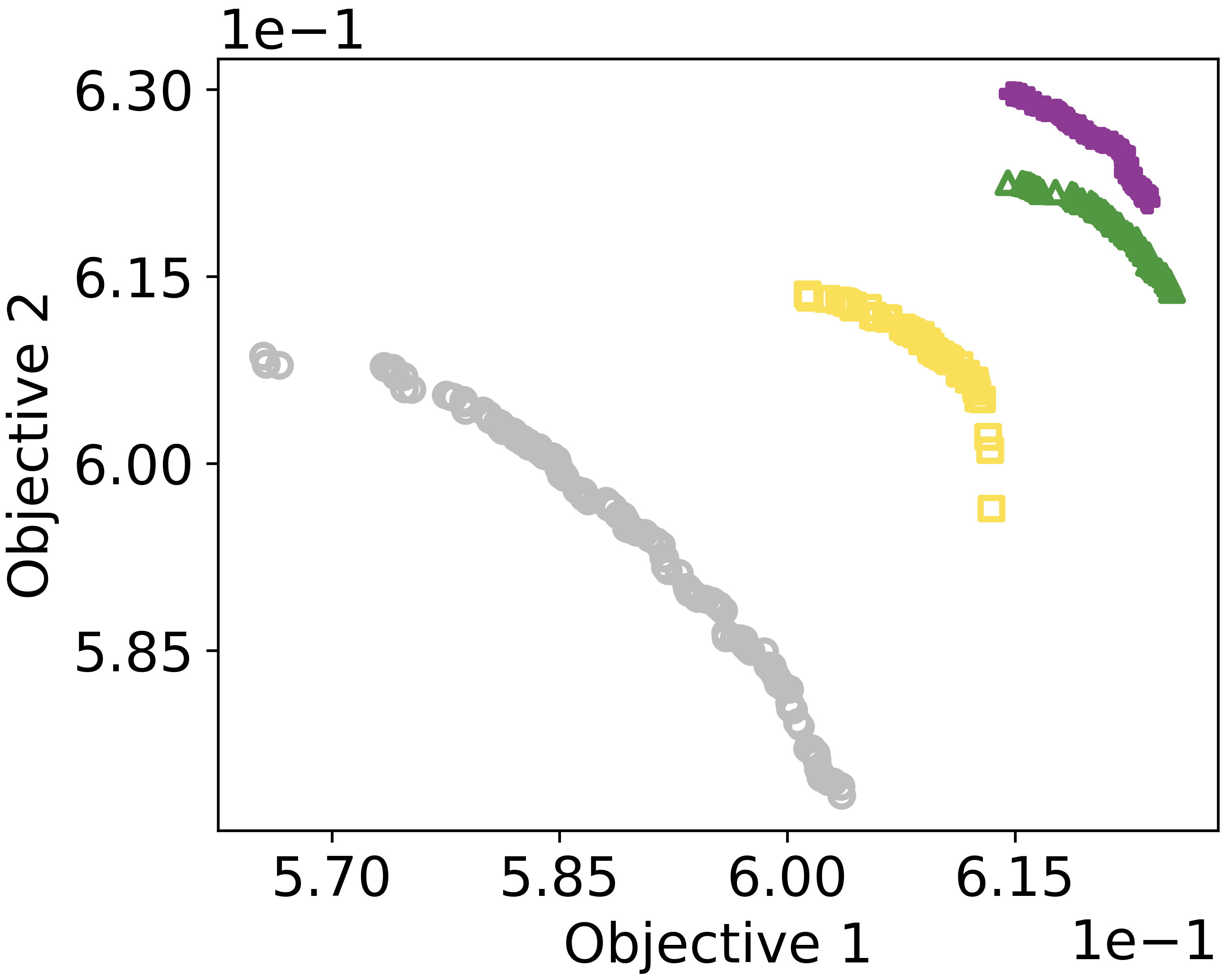} &
            \includegraphics[scale=0.31]{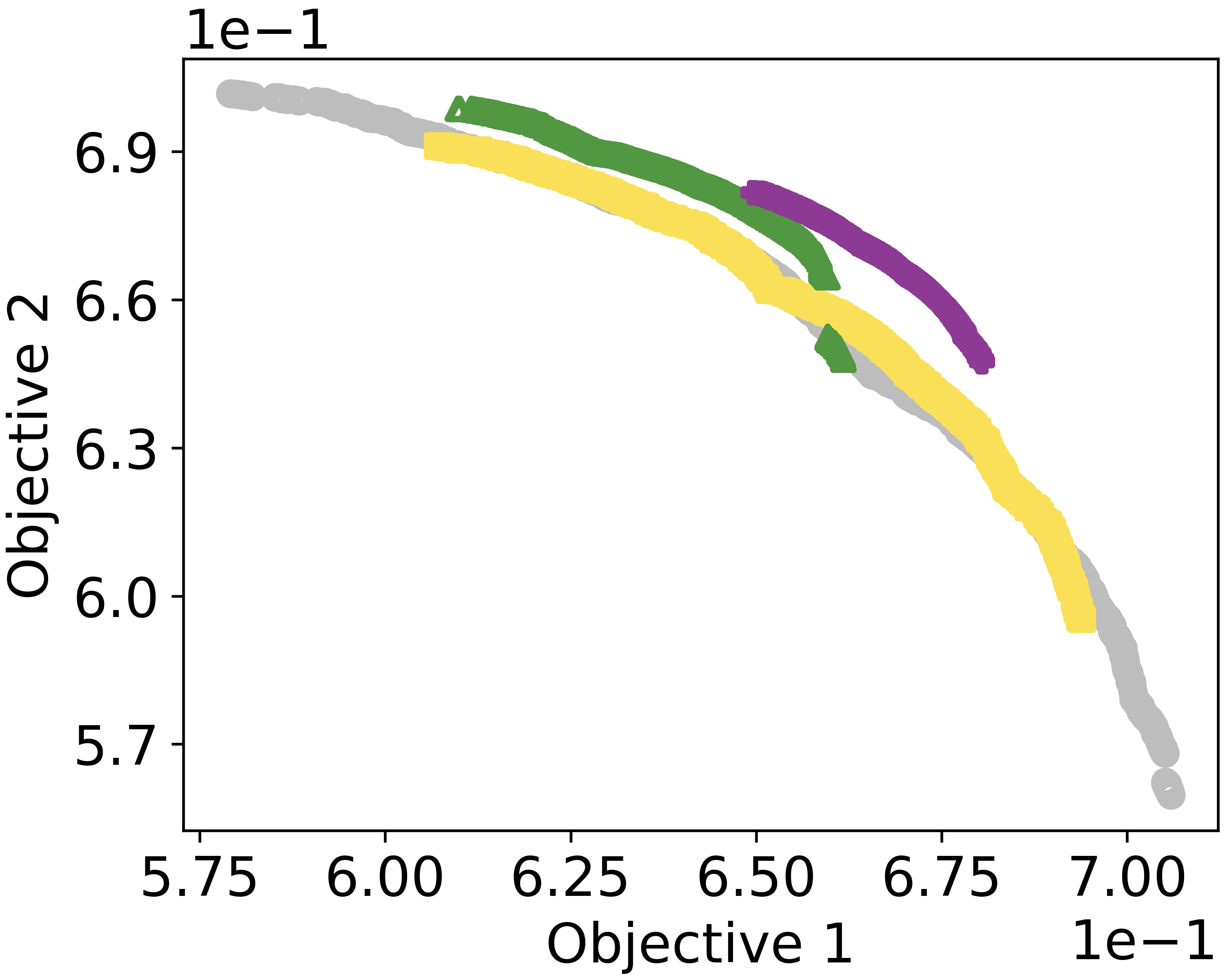}
             \\
			    (a) $10^5$ evaluations &
                (b) $10^7$ evaluations
		\end{tabular}
	\end{center}
        \vspace{-5pt}
	\caption{Solution sets obtained by the four algorithms in a representative run under two budgets of $10^5$ and $10^7$ evaluations on MONK with 5,000 decision variables}
	\label{Fig:MONK}
\end{figure}

Although the multi-objective NK-landscape problem (with $K=10$) has very rugged landscape, a similar performance pattern to that observed in the other problems can still be identified. However, unlike the previous two problems, SEMO exhibits a noticeably stronger scalability on this problem. As shown in Table~\ref{tab:MONK}, under the small budget, SEMO remains the best-performing algorithm on 100D, 500D and 1000D instances. A clear convergence gap between SEMO and the other algorithms only emerges when the problem size increases to $D=5000$. Figure~\ref{Fig:MONK} illustrates the solution sets obtained by the four algorithms on the 5000D instance under the two budget settings. However, despite being outperformed by the three MOEAs under the small budget, SEMO is able to catch up with the other algorithms and achieve better coverage under the large-budget setting.

\begin{table*}[!htbp]
\centering
\caption{HV results (mean and SD) of the four algorithms on the multi-objective NK-landscape (MONK) under the eight settings. The bold numbers for each algorithm denote the numbers of pairwise comparisons in which the algorithm performs statistically better, equally, or worse than its competitors, respectively.}
\vspace{-5pt}
\footnotesize
\begin{tabular}{clcccc}
\hline
Dimension &
  \multicolumn{1}{c}{Budget} &
  SEMO &
  NSGA-II &
  SMS-EMOA &
  MOEA/D \\ \hline

\multicolumn{1}{c|}{\multirow{4}{*}{100D}} &
  \multicolumn{1}{l|}{\multirow{2}{*}{$10^5$ evals}} &
  8.83e-02 (3.63e-03)&
  8.17e-02 (3.31e-03) &
  7.98e-02 (4.96e-03) &
  8.41e-02 (4.27e-03)\\
\multicolumn{1}{c|}{} &
  \multicolumn{1}{l|}{} &
  \textbf{3/0/0} &
  \textbf{0/2/1} &
  \textbf{0/1/2} &
  \textbf{1/1/1} \\ \cline{2-6}

\multicolumn{1}{c|}{} &
  \multicolumn{1}{l|}{\multirow{2}{*}{$10^7$ evals}} &
  8.87e-02 (3.43e-03) &
  9.39e-02 (4.22e-03) &
  9.26e-02 (3.64e-03) &
  9.23e-02 (5.17e-03) \\
\multicolumn{1}{c|}{} &
  \multicolumn{1}{l|}{} &
  \textbf{0/0/3} &
  \textbf{1/2/0} &
  \textbf{1/2/0} &
  \textbf{1/2/0} \\ \hline

\multicolumn{1}{c|}{\multirow{4}{*}{500D}} &
  \multicolumn{1}{l|}{\multirow{2}{*}{$10^5$ evals}} &
  5.07e-02 (1.54e-03) &
  4.24e-02 (2.29e-03) &
  4.30e-02 (1.59e-03) &
  4.54e-02 (1.69e-03) \\
\multicolumn{1}{c|}{} &
  \multicolumn{1}{l|}{} &
  \textbf{3/0/0} &
  \textbf{0/1/2} &
  \textbf{0/1/2} &
  \textbf{2/0/1} \\ \cline{2-6}

\multicolumn{1}{c|}{} &
  \multicolumn{1}{l|}{\multirow{2}{*}{$10^7$ evals}} &
  6.61e-02 (1.90e-03) &
  6.18e-02 (1.64e-03) &
  6.01e-02 (1.53e-03) &
  6.07e-02 (1.43e-03) \\
\multicolumn{1}{c|}{} &
  \multicolumn{1}{l|}{} &
  \textbf{3/0/0} &
  \textbf{2/0/1} &
  \textbf{0/1/2} &
  \textbf{0/1/2} \\ \hline

\multicolumn{1}{c|}{\multirow{4}{*}{1000D}} &
  \multicolumn{1}{l|}{\multirow{2}{*}{$10^5$ evals}} &
  3.82e-02 (8.69e-04) &
  3.32e-02 (2.06e-03) &
  3.44e-02 (1.13e-03) &
  3.69e-02 (1.06e-03) \\
\multicolumn{1}{c|}{} &
  \multicolumn{1}{l|}{} &
  \textbf{3/0/0} &
  \textbf{0/0/3} &
  \textbf{1/0/2} &
  \textbf{2/0/1} \\ \cline{2-6}

\multicolumn{1}{c|}{} &
  \multicolumn{1}{l|}{\multirow{2}{*}{$10^7$ evals}} &
   5.95e-02 (8.07e-04) &
  5.39e-02 (1.16e-03) &
  5.07e-02 (1.34e-03) &
  5.24e-02 (1.52e-03) \\
\multicolumn{1}{c|}{} &
  \multicolumn{1}{l|}{} &
  \textbf{3/0/0} &
  \textbf{2/0/1} &
  \textbf{0/0/3} &
  \textbf{1/0/2} \\ \hline

\multicolumn{1}{c|}{\multirow{4}{*}{5000D}} &
  \multicolumn{1}{l|}{\multirow{2}{*}{$10^5$ evals}} &
  1.32e-02 (5.70e-04) &
  1.45e-02 (7.43e-04) &
  1.81e-02 (4.90e-04) &
  1.88e-02 (4.41e-04) \\
\multicolumn{1}{c|}{} &
  \multicolumn{1}{l|}{} &
  \textbf{0/0/3} &
  \textbf{1/0/2} &
  \textbf{2/0/1} &
  \textbf{3/0/0} \\ \cline{2-6}

\multicolumn{1}{c|}{} &
  \multicolumn{1}{l|}{\multirow{2}{*}{$10^7$ evals}} &
  4.09e-02 (2.86e-04) &
  3.86e-02 (7.87e-04) &
  3.54e-02 (7.54e-04) &
  3.63e-02 (7.31e-04) \\
\multicolumn{1}{c|}{} &
  \multicolumn{1}{l|}{} &
  \textbf{3/0/0} &
  \textbf{2/0/1} &
  \textbf{0/0/3} &
  \textbf{1/0/2} \\ \hline

\end{tabular}
\label{tab:MONK}%
\end{table*} 

\subsection{On the Quadratic Assignment Problem}
On this problem, as shown in Table~\ref{tab:MOQAP}, the performance degradation pattern observed in the other problems has not yet been observed. SEMO consistently remains the best-performing algorithm across all tested settings. One possible reason is that the MOQAP has a rather rugged landscape, in which SEMO has shown clear advantage (as shown in \cite{Li2024}). Like the case on the NK-landscape, the pattern that MOQAP performs worse than the other three MOEAs becomes evident only when the problem size reaches $D = 5000$. For the MOQAP, instances of this scale are computationally prohibitive and require an excessive amount of running time\footnote{On the experimental platform used in this study, a single run of the MOQAP with $D=5000$ under the evaluation budget of $10^7$ requires more than ten days.}. As a result, we did not report experimental results on the 5000D instance. 

\begin{table*}[!htbp]
\centering
\caption{HV results (mean and SD) of the four algorithms on multi-objective QAP (MOQAP) under the eight settings. The bold numbers for each algorithm denote the numbers of pairwise comparisons in which the algorithm performs statistically better, equally, or worse than its competitors, respectively.}
\vspace{-5pt}
\footnotesize
\begin{tabular}{clcccc}
\hline
Dimension &
  \multicolumn{1}{c}{Budget} &
  SEMO &
  NSGA-II &
  SMS-EMOA &
  MOEA/D \\ \hline

\multicolumn{1}{c|}{\multirow{4}{*}{100D}} &
  \multicolumn{1}{l|}{\multirow{2}{*}{$10^5$ evals}} &
  3.42e+15 (1.10e+14)&
  2.51e+15 (1.56e+14) &
  2.60e+15 (1.44e+14) &
  2.59e+15 (1.62e+14)\\
\multicolumn{1}{c|}{} &
  \multicolumn{1}{l|}{} &
  \textbf{3/0/0} &
  \textbf{0/2/1} &
  \textbf{0/2/1} &
  \textbf{0/2/1} \\ \cline{2-6}

\multicolumn{1}{c|}{} &
  \multicolumn{1}{l|}{\multirow{2}{*}{$10^7$ evals}} &
  4.61e+15 (9.69e+13) &
  4.10e+15 (1.14e+14) &
  3.45e+15 (1.23e+14) &
  3.88e+15 (1.16e+14) \\
\multicolumn{1}{c|}{} &
  \multicolumn{1}{l|}{} &
  \textbf{3/0/0} &
  \textbf{2/0/1} &
  \textbf{0/0/3} &
  \textbf{1/0/2} \\ \hline

\multicolumn{1}{c|}{\multirow{4}{*}{500D}} &
  \multicolumn{1}{l|}{\multirow{2}{*}{$10^5$ evals}} &
  1.76e+17 (5.43e+15) &
  1.23e+17 (7.01e+15) &
  1.37e+17 (5.51e+15) &
  1.33e+17 (4.79e+15) \\
\multicolumn{1}{c|}{} &
  \multicolumn{1}{l|}{} &
  \textbf{3/0/0} &
  \textbf{0/0/3} &
  \textbf{2/0/1} &
  \textbf{1/0/2} \\ \cline{2-6}

\multicolumn{1}{c|}{} &
  \multicolumn{1}{l|}{\multirow{2}{*}{$10^7$ evals}} &
  4.17e+17 (3.81e+15)  &
  3.44e+17 (8.48e+15) &
  3.19e+17 (1.29e+16) &
  3.47e+17 (8.25e+15) \\
\multicolumn{1}{c|}{} &
  \multicolumn{1}{l|}{} &
  \textbf{3/0/0} &
  \textbf{1/1/1} &
  \textbf{0/0/3} &
  \textbf{1/1/1} \\ \hline

\multicolumn{1}{c|}{\multirow{4}{*}{1000D}} &
  \multicolumn{1}{l|}{\multirow{2}{*}{$10^5$ evals}} &
  8.70e+17 (3.05e+16) &
  6.06e+17 (3.50e+16) &
  6.78e+17 (3.04e+16) &
  6.56e+17 (2.85e+16) \\
\multicolumn{1}{c|}{} &
  \multicolumn{1}{l|}{} &
  \textbf{3/0/0} &
  \textbf{0/0/3} &
  \textbf{2/0/1} &
  \textbf{1/0/2} \\ \cline{2-6}

\multicolumn{1}{c|}{} &
  \multicolumn{1}{l|}{\multirow{2}{*}{$10^7$ evals}} &
  2.70e+18 (2.39e+16) &
  2.32e+18 (5.36e+16) &
  2.28e+18 (5.62e+16) &
  2.45e+18 (4.59e+16) \\
\multicolumn{1}{c|}{} &
  \multicolumn{1}{l|}{} &
  \textbf{3/0/0} &
  \textbf{1/0/2} &
  \textbf{0/0/3} &
  \textbf{2/0/1} \\ \hline

\end{tabular}
\label{tab:MOQAP}%
\end{table*}

\section{Incorporating Crossover into SEMO}

A key observation in previous section is that SEMO's convergence speed deteriorates as the problem size increases, compared with the other three MOEAs. A major difference of SEMO from the other algorithms is that SEMO only considers ``mutation'' as the variation operator, while the other three MOEAs consider both crossover and mutation. As shown in recent theoretical studies \cite{dang_crossover_2024,dang2023proof,opris2024many}, crossover can significantly accelerate the search for multi-objective combinatorial problems. Here, we incorporate crossover into SEMO, denoted by SEMOx, as a variant of the original SEMO. Specifically, two solutions are selected uniformly at random from the archive, and the same crossover in the other MOEAs is performed to generate two offspring. One of the offspring is then selected at random and subjected to the same mutation operator as in the original SEMO. Note that, as SEMO starts the search from a single point, the crossover operator is applied when there are at least two solutions in the archive.

\begin{figure}[tbp]
\vspace{-5pt}
	\begin{center}
		\hspace*{-0pt}
        \includegraphics[scale=0.28, trim=0 15 0 15]{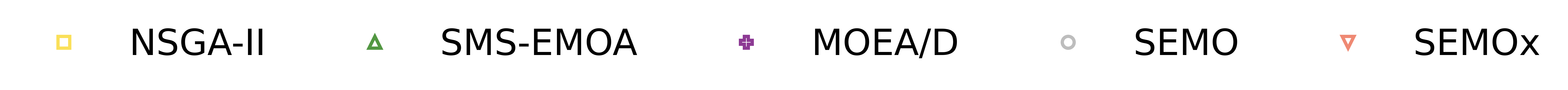}
        \vspace{-5pt}
        \begin{tabular}{@{}cc@{}}
			\includegraphics[scale=0.31]{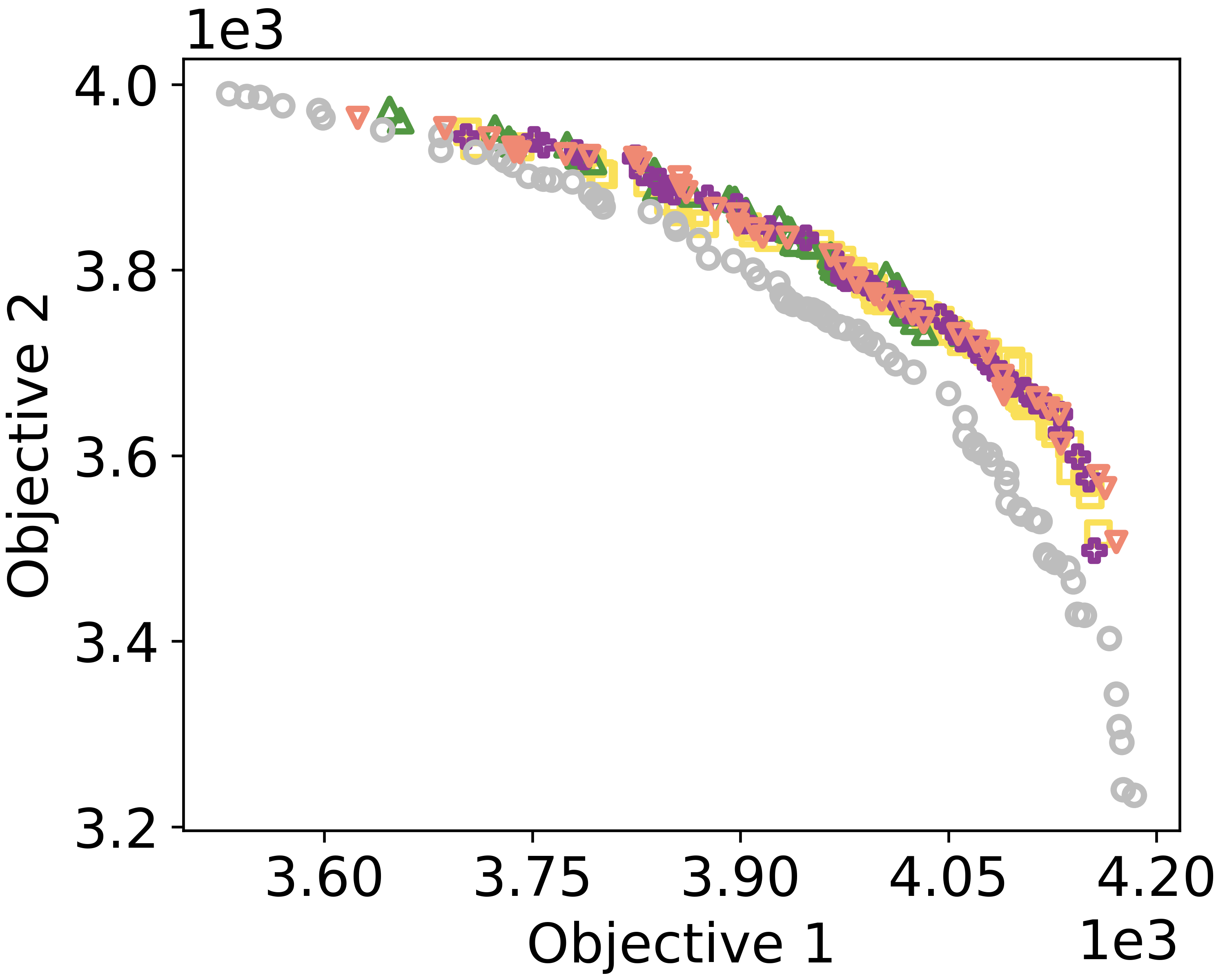} &
            \includegraphics[scale=0.31]{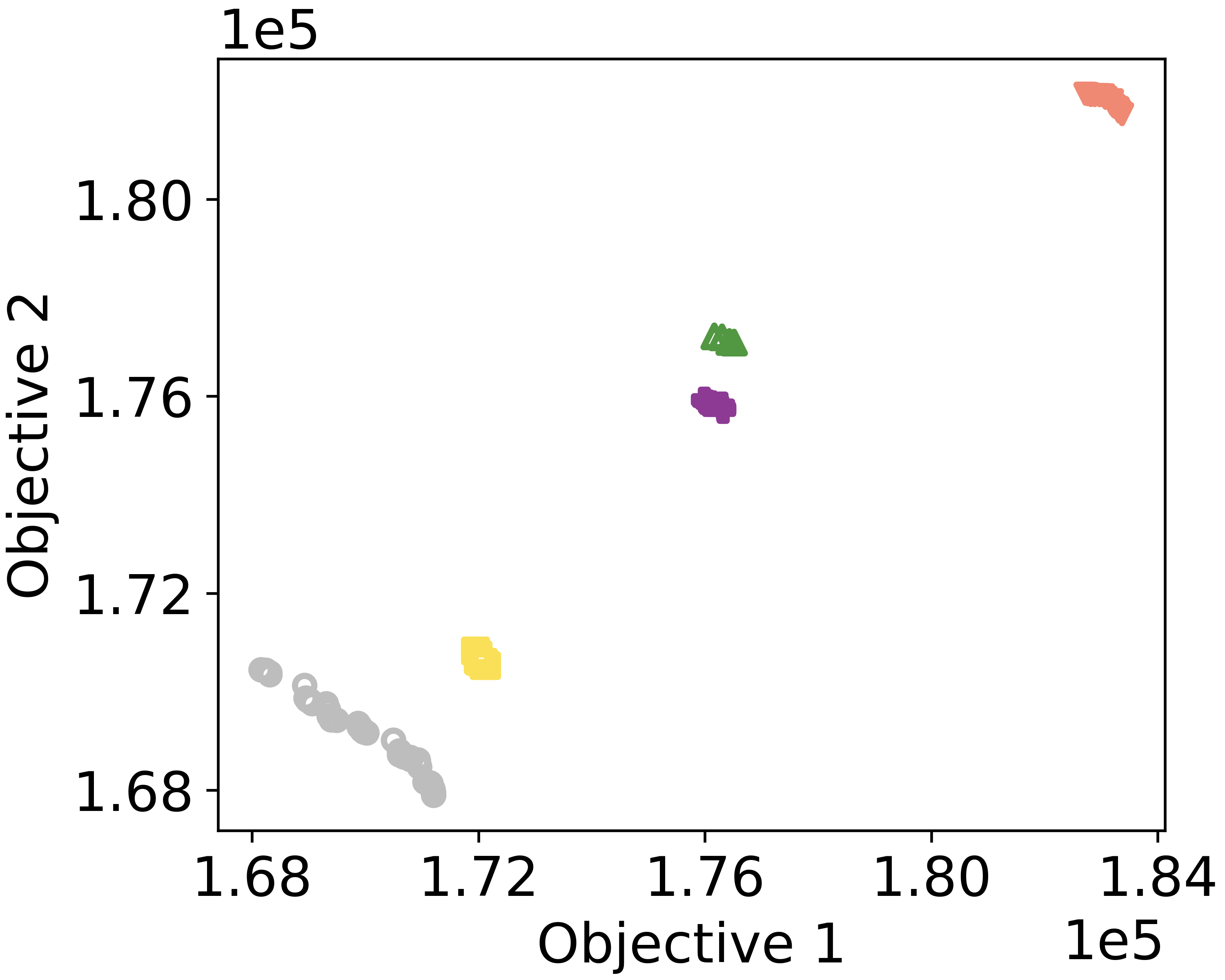}
             \\
			    (a) 100D &
                (b) 5000D
		\end{tabular}
	\end{center}
        \vspace{-5pt}
	\caption{Solution sets obtained by SEMO and SEMOx, along with the other three MOEAs, in a representative run under a budget of $10^5$ evaluations on MOKP with (a) 100 and (b) 5000 decision variables.}
	\label{Fig:SEMOx-1}
\end{figure}

Tables~\ref{tab:SEMO_SEMOx_1e5} presents the mean hypervolume (HV) values obtained by SEMOx, alongside by the original SEMO, on the four MOCOPs under the small evaluation budget. As can be seen in the table, incorporating crossover enables SEMO to work better on large-scale instances. Although SEMOx performs worse than the original SEMO on the 100D instances, they show clear advantage when the problem size increases, with a higher HV value on the 1000D and 5000D MOTSP, the 500D, 1000D and 5000D MOKP, the 5000D MONK, and the 500D and 1000D MOQAP. Figure \ref{Fig:SEMOx-1} shows the solution sets obtained by SEMOx and SEMO, alongside the other three MOEAs, on the 100D and 5000D MOKP problem instances. As seen, SEMOx exhibits substantially better convergence than the other algorithms on the large-scale instance.

\begin{table}[tbp]
\footnotesize
\renewcommand{\arraystretch}{1.2}
\centering
\caption{HV results (mean) of SEMO and SEMOx on the four multi-objective combinatorial optimisation problems under a budget of $10^{5}$ evaluations. For each problem and dimension, a larger HV value with statistical significance is highlighted in bold.}
\vspace{-5pt}
\label{tab:SEMO_SEMOx_1e5}
\begin{tabular}{llcccc}
\hline
Problem & Algorithm & 100D & 500D & 1000D & 5000D \\
\hline

\multirow{2}{*}{MOTSP}
 & SEMO  & \textbf{2.18e+03} & \textbf{2.63e+04} & 6.61e+04 & 1.82e+05 \\
 & SEMOx & 1.73e+03 & 2.30e+04 & \textbf{7.23e+04} & \textbf{5.62e+05} \\
\hline

\multirow{2}{*}{MOKP}
 & SEMO  & \textbf{2.51e+06} & 4.13e+07 & 1.50e+08 & 1.15e+09 \\
 & SEMOx & 2.50e+06 & \textbf{4.34e+07} & \textbf{1.72e+08} & \textbf{2.11e+09} \\
\hline

\multirow{2}{*}{MONK}
 & SEMO  & \textbf{8.83e-02} & \textbf{5.07e-02} & \textbf{3.82e-02} & 1.32e-02 \\
 & SEMOx & 8.61e-02 & 4.60e-02 & 3.74e-02 & \textbf{2.28e-02} \\
\hline

\multirow{2}{*}{MOQAP}
 & SEMO  & \textbf{3.42e+15} & 1.76e+17 & 8.70e+17 & N/A \\
 & SEMOx & 3.35e+15 & \textbf{2.53e+17} & \textbf{1.62e+18} & N/A \\
\hline
\end{tabular}
\end{table}

\begin{figure}[tbp]
\vspace{-5pt}
\centering
\begin{subfigure}{0.28\linewidth}
    \captionsetup{labelfont=normal,
  textfont=normal}
    \centering
    \includegraphics[width=\linewidth]{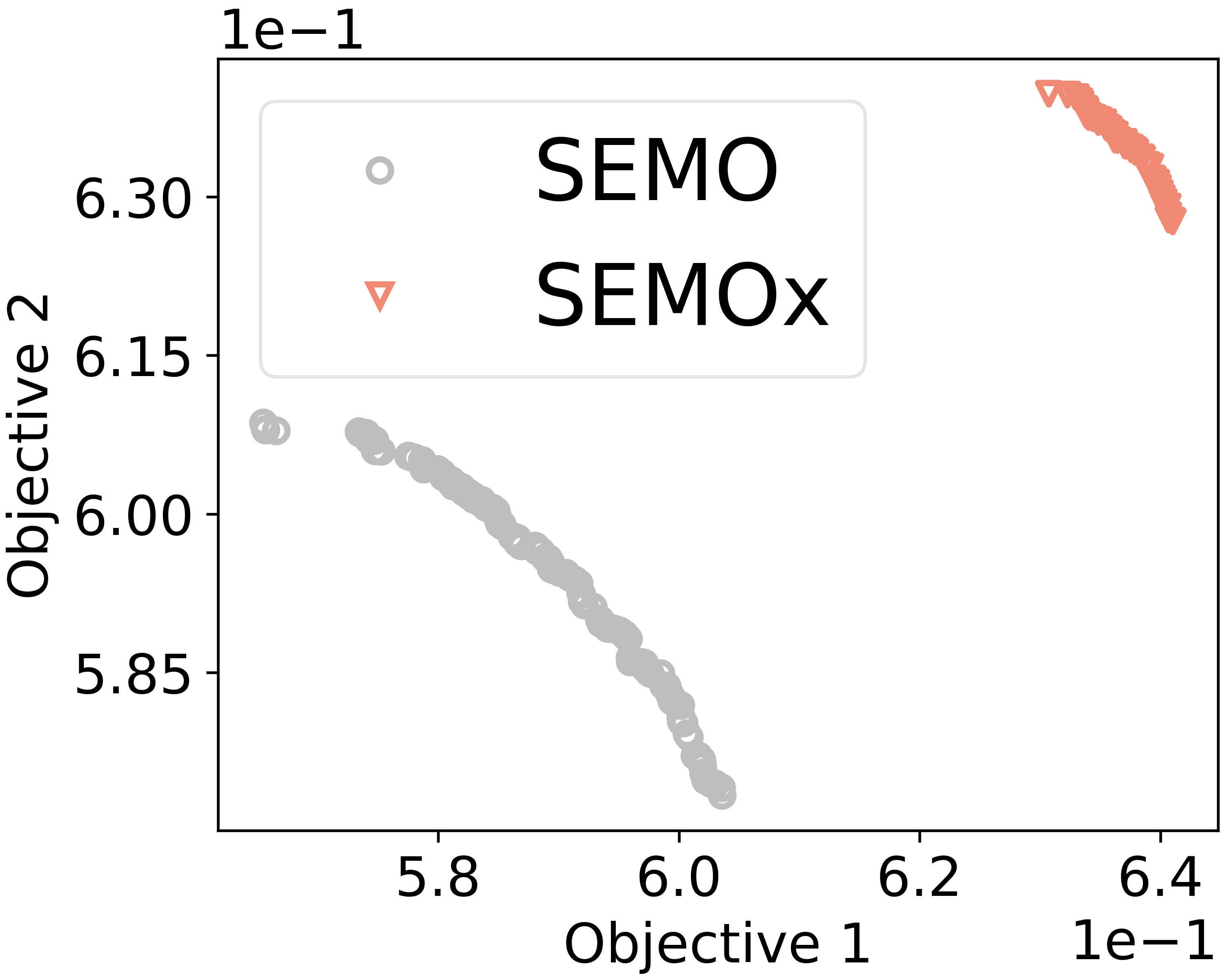}
    \caption{$10^5$ evaluations}
\end{subfigure}
\hfill
\begin{subfigure}{0.28\linewidth}
\captionsetup{labelfont=normal,
  textfont=normal}
    \centering
    \includegraphics[width=\linewidth]{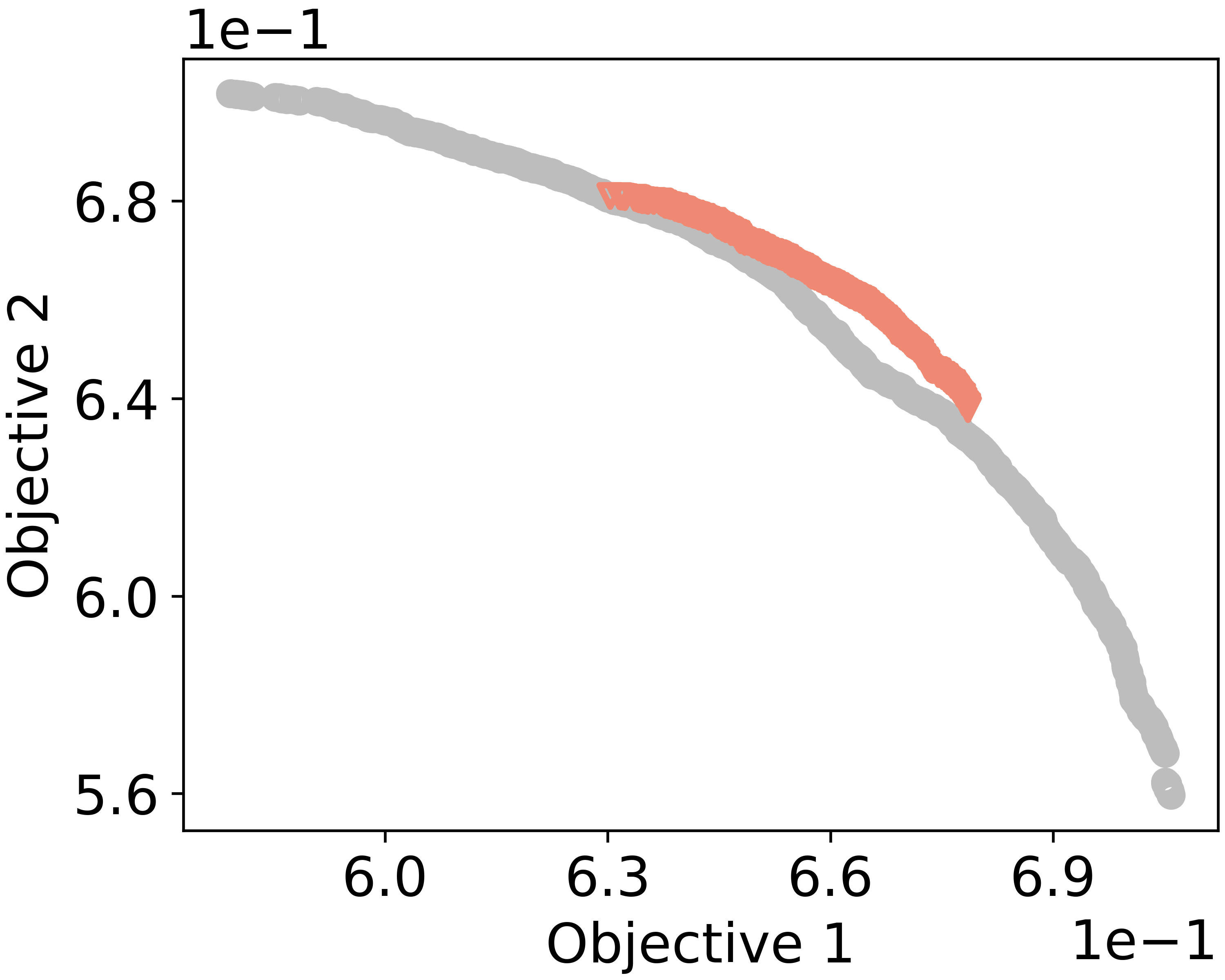}
    \caption{$10^7$ evaluations}
\end{subfigure}
\hfill
\begin{subfigure}{0.28\linewidth}
\captionsetup{labelfont=normal,
  textfont=normal}
    \centering
    \includegraphics[width=\linewidth]{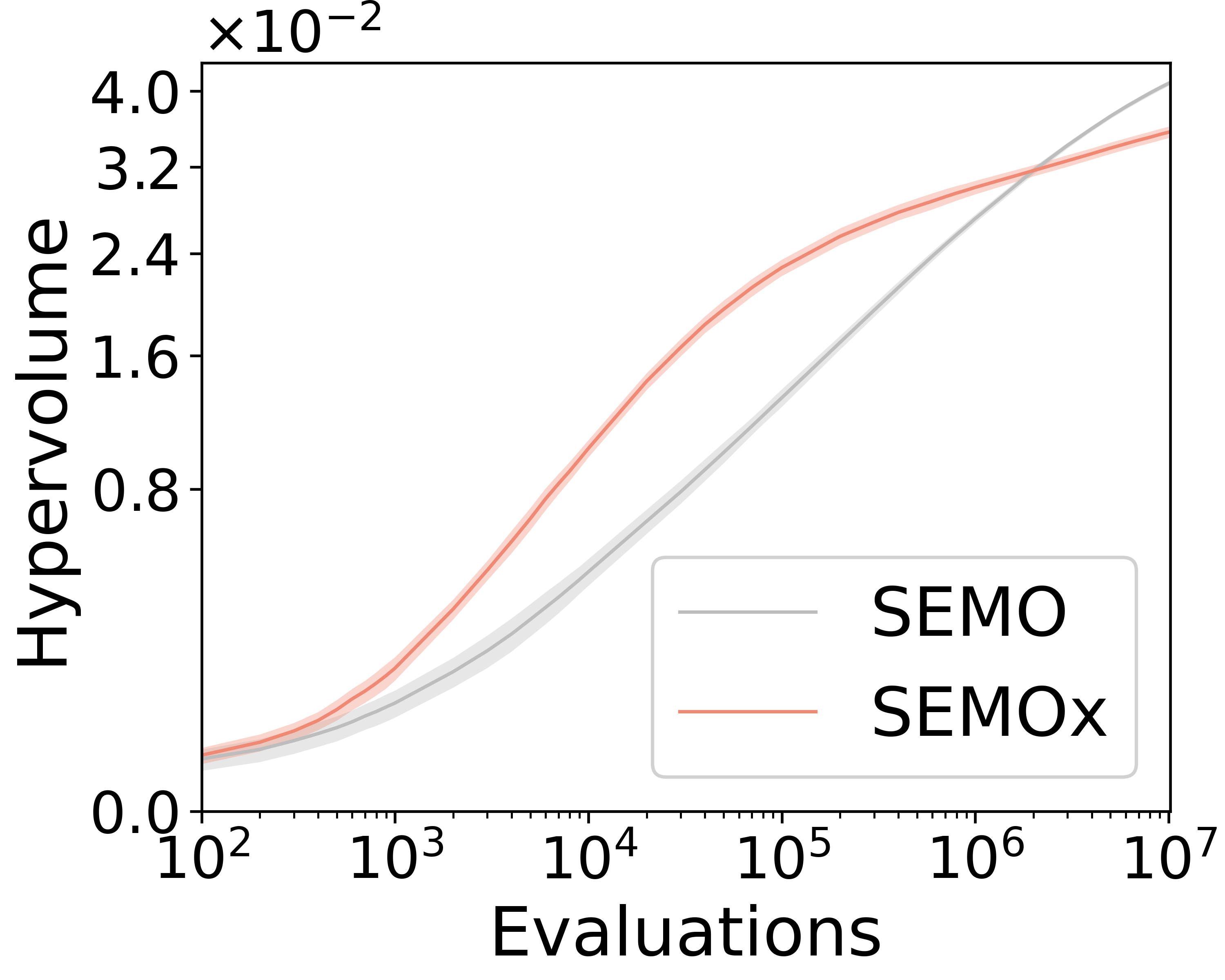}
    \caption{HV trajectory}
\end{subfigure}
\caption{Solution sets obtained by SEMO and SEMOx on the MONK with 5,000 decision variables under two evaluation budgets: 
(a) $10^5$ evaluations and (b) $10^7$ evaluations, alongside with (c) the HV trajectory of SEMO and SEMOx over the search process. 
In (c), the bold line and the shaded area represent the mean and standard deviation of the HV over 30 independent runs, respectively.}
\label{Fig:HV}
\end{figure}

However, as the same pattern observed in the three MOEAs, when a sufficiently large evaluation budget is available, the original SEMO can catch up. Table~\ref{tab:SEMO_SEMOx_1e7} presents the HV values obtained by the two SEMO versions under the evaluation budget of $10^{7}$. As can be seen in the table, the original SEMO achieves a higher HV value on most of the problem instances. Figure~\ref{Fig:HV} further illustrates this behaviour by presenting the solution sets obtained by SEMO and SEMOx on the MONK with 5,000 decision variables under two evaluation budgets, together with the corresponding HV trajectory over the search process. As shown in Figure~\ref{Fig:HV}(a), SEMOx demonstrates superior convergence under the budget of $10^5$ evaluations. However, in the large-budget setting in Figure~\ref{Fig:HV}(b), the original SEMO catches up and achieves a substantially large Pareto front. In fact, as can be seen from the HV trajectory in Figure~\ref{Fig:HV}(c), SEMOx maintains an advantage up to approximately $10^6$ evaluations, after which its HV value is surpassed by that of the original SEMO.

\begin{table}[tbp]
\footnotesize
\renewcommand{\arraystretch}{1.2}
\centering
\caption{HV results (mean) of SEMO and SEMOx on the four multi-objective combinatorial optimisation problems under a budget of $10^{7}$ evaluations. For each problem and dimension, a larger HV value with statistical significance is highlighted in bold.}
\vspace{-5pt}
\label{tab:SEMO_SEMOx_1e7}
\begin{tabular}{llcccc}
\hline
Problem & Algorithm & 100D & 500D & 1000D & 5000D \\
\hline

\multirow{2}{*}{MOTSP}
 & SEMO  & \textbf{2.69e+03} & \textbf{5.54e+04} & \textbf{1.90e+05} & \textbf{1.33e+06} \\
 & SEMOx & 2.42e+03 & 3.96e+04 & 1.30e+05 & 1.16e+06 \\
\hline

\multirow{2}{*}{MOKP}
 & SEMO  & 2.57e+06 & \textbf{5.01e+07} & \textbf{2.11e+08} & \textbf{3.03e+09} \\
 & SEMOx & \textbf{2.61e+06} & 4.93e+07 & 2.03e+08 & 3.00e+09 \\
\hline

\multirow{2}{*}{MONK}
 & SEMO  & 8.87e-02 & \textbf{6.61e-02} & \textbf{5.95e-02} & \textbf{4.09e-02} \\
 & SEMOx & \textbf{9.10e-02} & 6.20e-02 & 5.24e-02 & 3.56e-02 \\
\hline

\multirow{2}{*}{MOQAP}
 & SEMO  & \textbf{4.61e+15} & \textbf{4.17e+17} & 8.70e+17 & N/A \\
 & SEMOx & 4.14e+15 & 3.81e+17 & \textbf{2.74e+18} & N/A \\
\hline

\end{tabular}
\end{table}

Note that SEMOx starts from a single initial solution randomly drawn from the search space. As a result, the non-dominated solutions in the SEMOx archive tend to be similar, especially in the early stages of the search. Interestingly, this similarity can significantly speed up the search process. This observation is actually echoed by existing studies in the literature \cite{Ishibuchi2003,Ishibuchi2015,Ishibuchi2012}, which show that selecting similar parents for crossover can substantially enhance an algorithm's search ability on large-scale problems. 

However, this speedup may come at the cost of reduced population diversity, which is particularly important in multi-objective combinatorial optimisation problems, where the Pareto set typically consists of multiple disjoint Pareto optimal regions in the decision space. As such, SEMO without crossover may obtain a broader Pareto front when given sufficiently large budgets.

\section{Conclusions}
This paper presents an empirical study on the scalability of MOEAs for combinatorial optimisation problems, examining how differently algorithms behave as the problem size increases.
Based on the experimental results, several observations are obtained:
\begin{itemize}
    
    \item The convergence speed of SEMO degrades significantly as the problem size increases, compared with NSGA-II, SMS-EMOA and MOEA/D.
    
    \item An important reason for this behaviour is the absence of crossover. Incorporating crossover into SEMO can significant accelerate the search, suggesting that SEMO with cros-sover is preferable when operating under a tight budget. 
    
    \item With a sufficient computational budget, the original SEMO is able to catch up and often achieves a considerably better spread of its final solution set. However, for some small-scale problems (e.g., the 100D knapsack problem), SEMO may stagnate due to its reliance on a local search operator.

\end{itemize}

Future research may proceed along several directions. A natural extension is to study problems with more objectives, where the behaviours of MOEAs can be substantially different \cite{Allmendinger2022,liefooghe2023many}. Another direction is to perform detailed ablation studies that isolate the effects of crossover, mutation, and selection mechanisms to better understand the specific roles these components play in scalability, as in \cite{Aguirre2007, liefooghe2012dominance, Sagawa2017, bezerra2018large}. Finally, the findings of this study may help inform the design of more effective algorithms that strike a good balance between scalable convergence and solution diversity on large-scale combinatorial MOPs, for example, by adaptively switching the crossover operator on or off according to the state of the search.

\bibliographystyle{unsrt}
\bibliography{references}

@InProceedings{Li2023,
  author     = {Li, Miqing and Han, Xiaofeng and Chu, Xiaochen},
  booktitle  = {Proceedings of the Genetic and Evolutionary Computation Conference},
  title      = {{MOEAs} Are Stuck in a Different Area at a Time},
  year       = {2023},
  month      = jul,
  pages      = {303--311},
  publisher  = {ACM},
  series     = {GECCO ’23},
  collection = {GECCO ’23},
  doi        = {10.1145/3583131.3590447},
}

@InBook{Ochoa2023,
  author    = {Ochoa, Gabriela and Liefooghe, Arnaud and Lavinas, Yuri and Aranha, Claus},
  pages     = {211--226},
  publisher = {Springer Nature Switzerland},
  title     = {Decision/Objective Space Trajectory Networks for Multi-objective Combinatorial Optimisation},
  year      = {2023},
  isbn      = {9783031300356},
  booktitle = {Evolutionary Computation in Combinatorial Optimization},
  doi       = {10.1007/978-3-031-30035-6_14},
  issn      = {1611-3349},
}

@article{bezerra2018large,
  title={A large-scale experimental evaluation of high-performing multi-and many-objective evolutionary algorithms},
  author={Bezerra, Leonardo CT and L{\'o}pez-Ib{\'a}{\~n}ez, Manuel and St{\"u}tzle, Thomas},
  journal={Evolutionary computation},
  volume={26},
  number={4},
  pages={621--656},
  year={2018},
  publisher={MIT Press}
}

@inproceedings{liefooghe2023many,
  title={Many-objective (combinatorial) optimization is easy},
  author={Liefooghe, Arnaud and L{\'o}pez-Ib{\'a}{\~n}ez, Manuel},
  booktitle={Proceedings of the Genetic and Evolutionary Computation Conference},
  pages={704--712},
  year={2023}
}

@article{liefooghe2012dominance,
  title={On dominance-based multiobjective local search: design, implementation and experimental analysis on scheduling and traveling salesman problems},
  author={Liefooghe, Arnaud and Humeau, J{\'e}r{\'e}mie and Mesmoudi, Salma and Jourdan, Laetitia and Talbi, El-Ghazali},
  journal={Journal of Heuristics},
  volume={18},
  number={2},
  pages={317--352},
  year={2012},
  publisher={Springer}
}

@Article{Allmendinger2022,
  author    = {Allmendinger, Richard and Jaszkiewicz, Andrzej and Liefooghe, Arnaud and Tammer, Christiane},
  journal   = {Computers \& Operations Research},
  title     = {What if we increase the number of objectives? Theoretical and empirical implications for many-objective combinatorial optimization},
  year      = {2022},
  issn      = {0305-0548},
  month     = sep,
  pages     = {105857},
  volume    = {145},
  doi       = {10.1016/j.cor.2022.105857},
  publisher = {Elsevier BV},
}

@InProceedings{Sagawa2017,
  author    = {Sagawa, Miyako and Aguirre, Hernan and Daolio, Fabio and Liefooghe, Arnaud and Derbel, Bilel and Verel, Sebastien and Tanaka, Kiyoshi},
  booktitle = {2017 6th IIAI International Congress on Advanced Applied Informatics (IIAI-AAI)},
  title     = {Learning Variable Importance to Guide Recombination on Many-Objective Optimization},
  year      = {2017},
  month     = jul,
  pages     = {874--879},
  publisher = {IEEE},
  doi       = {10.1109/iiai-aai.2017.158},
}

@InProceedings{Fieldsend2025,
  author     = {Fieldsend, Jonathan and Liefooghe, Arnaud and Malan, Katherine and Verel, Sébastien},
  booktitle  = {Proceedings of the Genetic and Evolutionary Computation Conference},
  title      = {Local Optima Networks for Constrained Search Spaces},
  year       = {2025},
  month      = jul,
  pages      = {204--212},
  publisher  = {ACM},
  series     = {GECCO ’25},
  collection = {GECCO ’25},
  doi        = {10.1145/3712256.3726303},
}

@article{dang_crossover_2024,
	title = {Crossover can guarantee exponential speed-ups in evolutionary multi-objective optimisation},
	volume = {330},
	issn = {0004-3702},
	doi = {10.1016/j.artint.2024.104098},
	journal = {Artificial Intelligence},
	author = {Dang, Duc-Cuong and Opris, Andre and Sudholt, Dirk},
	month = may,
	year = {2024},
	pages = {104098},
}

@article{opris2024many,
  title={A Many Objective Problem Where Crossover is Provably Indispensable},
  author={Opris, Andre},
  journal={arXiv preprint arXiv:2412.18375},
  year={2024}
}

@inproceedings{dang2023proof,
  title={A proof that using crossover can guarantee exponential speed-ups in evolutionary multi-objective optimisation},
  author={Dang, Duc-Cuong and Opris, Andre and Salehi, Bahare and Sudholt, Dirk},
  booktitle={AAAI Conference on Artificial Intelligence},
  volume={37},
  number={10},
  pages={12390--12398},
  year={2023}
}

@InBook{Ishibuchi2003,
  author    = {Ishibuchi, Hisao and Shibata, Youhei},
  pages     = {433--447},
  publisher = {Springer Berlin Heidelberg},
  title     = {An Empirical Study on the Effect of Mating Restriction on the Search Ability of {EMO} Algorithms},
  year      = {2003},
  isbn      = {9783540369707},
  booktitle = {Evolutionary Multi-Criterion Optimization},
  doi       = {10.1007/3-540-36970-8_31},
  issn      = {0302-9743},
}

@InBook{Ishibuchi2012,
  author    = {Ishibuchi, Hisao and Akedo, Naoya and Nojima, Yusuke},
  pages     = {132--142},
  publisher = {Springer Berlin Heidelberg},
  title     = {Recombination of Similar Parents in {SMS-EMOA} on Many-Objective 0/1 Knapsack Problems},
  year      = {2012},
  isbn      = {9783642329647},
  booktitle = {Parallel Problem Solving from Nature - PPSN XII},
  doi       = {10.1007/978-3-642-32964-7_14},
  issn      = {1611-3349},
}

@Article{Behmanesh2020,
  author    = {Behmanesh, Reza and Rahimi, Iman and Gandomi, Amir H.},
  journal   = {Archives of Computational Methods in Engineering},
  title     = {Evolutionary Many-Objective Algorithms for Combinatorial Optimization Problems: A Comparative Study},
  year      = {2020},
  issn      = {1886-1784},
  month     = mar,
  number    = {2},
  pages     = {673--688},
  volume    = {28},
  doi       = {10.1007/s11831-020-09415-3},
  publisher = {Springer Science and Business Media LLC},
}

@InBook{Liefooghe2020,
  author    = {Liefooghe, Arnaud and Verel, Sébastien and Derbel, Bilel and Aguirre, Hernan and Tanaka, Kiyoshi},
  pages     = {33--47},
  publisher = {Springer International Publishing},
  title     = {Dominance, Indicator and Decomposition Based Search for Multi-objective {QAP}: Landscape Analysis and Automated Algorithm Selection},
  year      = {2020},
  isbn      = {9783030581121},
  booktitle = {Parallel Problem Solving from Nature – PPSN XVI},
  doi       = {10.1007/978-3-030-58112-1_3},
  issn      = {1611-3349},
}

@article{figueira2017easy,
  title={Easy to say they are hard, but hard to see they are easy—towards a categorization of tractable multiobjective combinatorial optimization problems},
  author={Figueira, Jos{\'e} Rui and Fonseca, Carlos M and Halffmann, Pascal and Klamroth, Kathrin and Paquete, Lu{\'\i}s and Ruzika, Stefan and Schulze, Britta and Stiglmayr, Michael and Willems, David},
  journal={Journal of Multi-Criteria Decision Analysis},
  volume={24},
  number={1-2},
  pages={82--98},
  year={2017},
  publisher={Wiley Online Library}
}

@inproceedings{Liang2025,
author = {Liang, Zimin and Li, Miqing},
title = {On the problem characteristics of multi-objective pseudo-{Boolean} functions in runtime analysis},
year = {2025},
booktitle = {FOGA},
pages = {166–177},
numpages = {12},
}

@Article{Holm1979,
  author    = {Sture Holm},
  journal   = {Scandinavian Journal of Statistics},
  title     = {A Simple Sequentially Rejective Multiple Test Procedure},
  year      = {1979},
  issn      = {03036898, 14679469},
  number    = {2},
  pages     = {65--70},
  volume    = {6},
  publisher = {[Board of the Foundation of the Scandinavian Journal of Statistics, Wiley]},
  url       = {http://www.jstor.org/stable/4615733},
  urldate   = {2026-01-12},
}

@Article{Friedman1940,
  author    = {Friedman, Milton},
  journal   = {The Annals of Mathematical Statistics},
  title     = {A Comparison of Alternative Tests of Significance for the Problem of $m$ Rankings},
  year      = {1940},
  issn      = {0003-4851},
  month     = mar,
  number    = {1},
  pages     = {86--92},
  volume    = {11},
  doi       = {10.1214/aoms/1177731944},
  publisher = {Institute of Mathematical Statistics},
}

@InBook{Haynes2013,
  author    = {Haynes, Winston},
  pages     = {2354--2355},
  publisher = {Springer New York},
  title     = {Wilcoxon Rank Sum Test},
  year      = {2013},
  isbn      = {9781441998637},
  booktitle = {Encyclopedia of Systems Biology},
  doi       = {10.1007/978-1-4419-9863-7_1185},
}

@Book{Eiben2015,
  author    = {Eiben, A.E. and Smith, J.E.},
  publisher = {Springer Berlin Heidelberg},
  title     = {Introduction to Evolutionary Computing},
  year      = {2015},
  isbn      = {9783662448748},
  doi       = {10.1007/978-3-662-44874-8},
  issn      = {1619-7127},
  journal   = {Natural Computing Series},
}

@Article{Shang2021,
  author    = {Shang, Ke and Ishibuchi, Hisao and He, Linjun and Pang, Lie Meng},
  journal   = {IEEE Transactions on Evolutionary Computation},
  title     = {A Survey on the Hypervolume Indicator in Evolutionary Multiobjective Optimization},
  year      = {2021},
  issn      = {1941-0026},
  month     = feb,
  number    = {1},
  pages     = {1--20},
  volume    = {25},
  doi       = {10.1109/tevc.2020.3013290},
  publisher = {Institute of Electrical and Electronics Engineers (IEEE)},
}

@Article{Li2022,
  author    = {Li, Miqing and Chen, Tao and Yao, Xin},
  journal   = {IEEE Transactions on Software Engineering},
  title     = {How to Evaluate Solutions in Pareto-Based Search-Based Software Engineering: A Critical Review and Methodological Guidance},
  year      = {2022},
  issn      = {2326-3881},
  month     = may,
  number    = {5},
  pages     = {1771--1799},
  volume    = {48},
  doi       = {10.1109/tse.2020.3036108},
  publisher = {Institute of Electrical and Electronics Engineers (IEEE)},
}

@InProceedings{Daolio2015,
  author     = {Daolio, Fabio and Liefooghe, Arnaud and Verel, Sébastien and Aguirre, Hernán and Tanaka, Kiyoshi},
  booktitle  = {Proceedings of the 2015 Annual Conference on Genetic and Evolutionary Computation},
  title      = {Global vs Local Search on Multi-objective NK-Landscapes: Contrasting the Impact of Problem Features},
  year       = {2015},
  month      = jul,
  pages      = {369--376},
  publisher  = {ACM},
  series     = {GECCO ’15},
  collection = {GECCO ’15},
  doi        = {10.1145/2739480.2754745},
}

@Article{Taillard1995,
  author    = {Taillard, Éric D.},
  journal   = {Location Science},
  title     = {Comparison of iterative searches for the quadratic assignment problem},
  year      = {1995},
  issn      = {0966-8349},
  month     = aug,
  number    = {2},
  pages     = {87--105},
  volume    = {3},
  doi       = {10.1016/0966-8349(95)00008-6},
  publisher = {Elsevier BV},
}

@InProceedings{Corne2007,
  author     = {Corne, David W. and Knowles, Joshua D.},
  booktitle  = {Proceedings of the 9th annual conference on Genetic and evolutionary computation},
  title      = {Techniques for highly multiobjective optimisation: some nondominated points are better than others},
  year       = {2007},
  month      = jul,
  pages      = {773--780},
  publisher  = {ACM},
  series     = {GECCO07},
  collection = {GECCO07},
  doi        = {10.1145/1276958.1277115},
}

@InProceedings{Tanabe2017,
  author     = {Tanabe, Ryoji and Oyama, Akira},
  booktitle  = {Proceedings of the Genetic and Evolutionary Computation Conference},
  title      = {Benchmarking {MOEAs} for multi- and many-objective optimization using an unbounded external archive},
  year       = {2017},
  month      = jul,
  pages      = {633--640},
  publisher  = {ACM},
  series     = {GECCO ’17},
  collection = {GECCO ’17},
  doi        = {10.1145/3071178.3079192},
}

@Article{Li2024a,
  author    = {Li, Miqing and López-Ibáñez, Manuel and Yao, Xin},
  journal   = {IEEE Transactions on Evolutionary Computation},
  title     = {Multi-Objective Archiving},
  year      = {2024},
  issn      = {1941-0026},
  month     = jun,
  number    = {3},
  pages     = {696--717},
  volume    = {28},
  doi       = {10.1109/tevc.2023.3314152},
  publisher = {Institute of Electrical and Electronics Engineers (IEEE)},
}

@InBook{Li2019,
  author    = {Li, Miqing and Yao, Xin},
  pages     = {15--26},
  publisher = {Springer International Publishing},
  title     = {An Empirical Investigation of the Optimality and Monotonicity Properties of Multiobjective Archiving Methods},
  year      = {2019},
  isbn      = {9783030125981},
  booktitle = {Evolutionary Multi-Criterion Optimization},
  doi       = {10.1007/978-3-030-12598-1_2},
  issn      = {1611-3349},
}

@Article{Li2019a,
  Title                    = {Quality evaluation of solution sets in multiobjective optimisation: A survey},
  Author                   = {Li, Miqing and Yao, Xin},
  Journal                  = {ACM Computing Surveys},
  Year                     = {2019},
  Number                   = {2},
}

@InProceedings{Bian2018,
  author     = {Bian, Chao and Qian, Chao and Tang, Ke},
  booktitle  = {Proceedings of the Twenty-Seventh International Joint Conference on Artificial Intelligence},
  title      = {A General Approach to Running Time Analysis of Multi-objective Evolutionary Algorithms},
  year       = {2018},
  month      = jul,
  pages      = {1405--1411},
  publisher  = {International Joint Conferences on Artificial Intelligence Organization},
  series     = {IJCAI-2018},
  collection = {IJCAI-2018},
  doi        = {10.24963/ijcai.2018/195},
}

@InProceedings{Doerr2021,
  author     = {Doerr, Benjamin and Zheng, Weijie},
  booktitle  = {Proceedings of the Genetic and Evolutionary Computation Conference Companion},
  title      = {Theoretical analyses of multi-objective evolutionary algorithms on multi-modal objectives: (hot-off-the-press track at GECCO 2021)},
  year       = {2021},
  month      = jul,
  pages      = {25--26},
  publisher  = {ACM},
  series     = {GECCO ’21},
  collection = {GECCO ’21},
  doi        = {10.1145/3449726.3462719},
}

@InProceedings{Giel2006,
  author     = {Giel, Oliver and Lehre, Per Kristian},
  booktitle  = {Proceedings of the 8th annual conference on Genetic and evolutionary computation},
  title      = {On the effect of populations in evolutionary multi-objective optimization},
  year       = {2006},
  month      = jul,
  pages      = {651--658},
  publisher  = {ACM},
  series     = {GECCO06},
  collection = {GECCO06},
  doi        = {10.1145/1143997.1144114},
}

@Article{Osuna2018,
  author    = {Osuna, Edgar Covantes and Gao, Wanru and Neumann, Frank and Sudholt, Dirk},
  title     = {Design and Analysis of Diversity-Based Parent Selection Schemes for Speeding Up Evolutionary Multi-objective Optimisation},
  year      = {2018},
  copyright = {arXiv.org perpetual, non-exclusive license},
  doi       = {10.48550/ARXIV.1805.01221},
  publisher = {arXiv},
}

@Article{Qian2013,
  author    = {Qian, Chao and Yu, Yang and Zhou, Zhi-Hua},
  journal   = {Artificial Intelligence},
  title     = {An analysis on recombination in multi-objective evolutionary optimization},
  year      = {2013},
  issn      = {0004-3702},
  month     = nov,
  pages     = {99--119},
  volume    = {204},
  doi       = {10.1016/j.artint.2013.09.002},
  publisher = {Elsevier BV},
}

@inproceedings{liang2026random,
  title={Random is faster than systematic in multi-objective local search},
  author={Liang, Zimin and Li, Miqing},
  booktitle={Proceedings of the AAAI Conference on Artificial Intelligence},
  year={2026}
}

@Article{Jaszkiewicz2002,
  author    = {Jaszkiewicz, A.},
  journal   = {IEEE Transactions on Evolutionary Computation},
  title     = {On the performance of multiple-objective genetic local search on the 0/1 knapsack problem - a comparative experiment},
  year      = {2002},
  issn      = {1089-778X},
  month     = aug,
  number    = {4},
  pages     = {402--412},
  volume    = {6},
  doi       = {10.1109/tevc.2002.802873},
  publisher = {Institute of Electrical and Electronics Engineers (IEEE)},
}

@InProceedings{Merz,
  author     = {Merz, P. and Freisleben, B.},
  booktitle  = {1998 IEEE International Conference on Evolutionary Computation Proceedings. IEEE World Congress on Computational Intelligence (Cat. No.98TH8360)},
  title      = {On the effectiveness of evolutionary search in high-dimensional {NK-landscapes}},
  pages      = {741--745},
  publisher  = {IEEE},
  series     = {ICEC-98},
  collection = {ICEC-98},
  doi        = {10.1109/icec.1998.700144},
}

@Article{Psychas2015,
  author    = {Psychas, Iraklis-Dimitrios and Delimpasi, Eleni and Marinakis, Yannis},
  journal   = {Expert Systems with Applications},
  title     = {Hybrid evolutionary algorithms for the Multiobjective Traveling Salesman Problem},
  year      = {2015},
  issn      = {0957-4174},
  month     = dec,
  number    = {22},
  pages     = {8956--8970},
  volume    = {42},
  doi       = {10.1016/j.eswa.2015.07.051},
  publisher = {Elsevier BV},
}

@InProceedings{Aguirre2004,
  author     = {Aguirre, H.E. and Tanaka, K.},
  booktitle  = {Proceedings of the 2004 Congress on Evolutionary Computation (IEEE Cat. No.04TH8753)},
  title      = {Effects of elitism and population climbing on multiobjective {MNK-landscapes}},
  year      = {2004},
  pages      = {449--456},
  publisher  = {IEEE},
  series     = {CEC-04},
  collection = {CEC-04},
  doi        = {10.1109/cec.2004.1330891},
}

@InBook{Borges2002,
  author    = {Borges, Pedro Castro and Hansen, Michael Pilegaard},
  pages     = {129--150},
  publisher = {Springer US},
  title     = {A Study of Global Convexity for a Multiple Objective Travelling Salesman Problem},
  year      = {2002},
  isbn      = {9781461515074},
  booktitle = {Essays and Surveys in Metaheuristics},
  doi       = {10.1007/978-1-4615-1507-4_6},
  issn      = {1387-666X},
}

@InBook{Knowles2003,
  author    = {Knowles, Joshua and Corne, David},
  pages     = {295--310},
  publisher = {Springer Berlin Heidelberg},
  title     = {Instance Generators and Test Suites for the Multiobjective Quadratic Assignment Problem},
  year      = {2003},
  isbn      = {9783540369707},
  booktitle = {Evolutionary Multi-Criterion Optimization},
  doi       = {10.1007/3-540-36970-8_21},
  issn      = {0302-9743},
}

@Article{Shim2011,
  author    = {Shim, Vui Ann and Tan, Kay Chen and Chia, Jun Yong and Chong, Jin Kiat},
  journal   = {Flexible Services and Manufacturing Journal},
  title     = {Evolutionary algorithms for solving multi-objective travelling salesman problem},
  year      = {2011},
  issn      = {1936-6590},
  month     = jun,
  number    = {2},
  pages     = {207--241},
  volume    = {23},
  doi       = {10.1007/s10696-011-9099-y},
  publisher = {Springer Science and Business Media LLC},
}

@Article{Zitzler1999,
  author    = {Zitzler, E. and Thiele, L.},
  journal   = {IEEE Transactions on Evolutionary Computation},
  title     = {Multiobjective evolutionary algorithms: a comparative case study and the strength Pareto approach},
  year      = {1999},
  issn      = {1089-778X},
  number    = {4},
  pages     = {257--271},
  volume    = {3},
  doi       = {10.1109/4235.797969},
  publisher = {Institute of Electrical and Electronics Engineers (IEEE)},
}

@InProceedings{Sastry,
  author    = {Sastry, K. and Goldberg, D.E. and Pelikan, M.},
  booktitle = {2005 IEEE Congress on Evolutionary Computation},
  title     = {Limits of Scalability of Multiobjective Estimation of Distribution Algorithms},
  pages     = {2217--2224},
  publisher = {IEEE},
  volume    = {3},
  doi       = {10.1109/cec.2005.1554970},
}

@Article{Hong2019,
  author    = {Hong, Wenjing and Tang, Ke and Zhou, Aimin and Ishibuchi, Hisao and Yao, Xin},
  journal   = {IEEE Transactions on Evolutionary Computation},
  title     = {A Scalable Indicator-Based Evolutionary Algorithm for Large-Scale Multiobjective Optimization},
  year      = {2019},
  issn      = {1941-0026},
  month     = jun,
  number    = {3},
  pages     = {525--537},
  volume    = {23},
  doi       = {10.1109/tevc.2018.2881153},
  publisher = {Institute of Electrical and Electronics Engineers (IEEE)},
}

@InBook{MiguelAntonio2016,
  author    = {Miguel Antonio, Luis and Coello Coello, Carlos A.},
  pages     = {525--534},
  publisher = {Springer International Publishing},
  title     = {Decomposition-Based Approach for Solving Large Scale Multi-objective Problems},
  year      = {2016},
  isbn      = {9783319458236},
  booktitle = {Parallel Problem Solving from Nature – PPSN XIV},
  doi       = {10.1007/978-3-319-45823-6_49},
  issn      = {1611-3349},
}

@InBook{Castillo2003,
  author    = {Castillo, P. A. and Arenas, M. G. and Castillo-Valdivieso, J. J. and Merelo, J. J. and Prieto, A. and Romero, G.},
  pages     = {43--52},
  publisher = {Springer London},
  title     = {Artificial Neural Networks Design using Evolutionary Algorithms},
  year      = {2003},
  isbn      = {9781447137443},
  booktitle = {Advances in Soft Computing},
  doi       = {10.1007/978-1-4471-3744-3_5},
}

@Article{Hierons2016SIP,
  Title                    = {{SIP}: Optimal Product Selection from Feature Models Using Many-Objective Evolutionary Optimization},
  Author                   = {Hierons, Robert M. and Li, Miqing and Liu, Xiaohui and Segura, Sergio and Zheng, Wei},
  Journal                  = {ACM Transactions on Software Engineering and Methodology},
  Year                     = {2016},
  Number                   = {2},
  Pages                    = {1-39},
  Volume                   = {25}
}

@InProceedings{Li2024,
  author     = {Li, Miqing and Han, Xiaofeng and Chu, Xiaochen and Liang, Zimin},
  booktitle  = {Proceedings of the Genetic and Evolutionary Computation Conference},
  title      = {Empirical Comparison between {MOEAs} and Local Search on Multi-Objective Combinatorial Optimisation Problems},
  year       = {2024},
  month      = jul,
  pages      = {547--556},
  publisher  = {ACM},
  series     = {GECCO ’24},
  collection = {GECCO ’24},
  doi        = {10.1145/3638529.3654077},
}

@Article{Jin2008,
  author    = {Yaochu Jin and Sendhoff, B.},
  journal   = {IEEE Transactions on Systems, Man, and Cybernetics, Part C (Applications and Reviews)},
  title     = {Pareto-Based Multiobjective Machine Learning: An Overview and Case Studies},
  year      = {2008},
  issn      = {1094-6977},
  month     = may,
  number    = {3},
  pages     = {397--415},
  volume    = {38},
  doi       = {10.1109/tsmcc.2008.919172},
  publisher = {Institute of Electrical and Electronics Engineers (IEEE)},
}

@Article{Xiang2017,
  author    = {Xiang, Yi and Zhou, Yuren and Zheng, Zibin and Li, Miqing},
  journal   = {ACM Transactions on Software Engineering and Methodology},
  title     = {Configuring Software Product Lines by Combining Many-Objective Optimization and {SAT} Solvers},
  year      = {2017},
  issn      = {1557-7392},
  month     = oct,
  number    = {4},
  pages     = {1--46},
  volume    = {26},
  doi       = {10.1145/3176644},
  publisher = {Association for Computing Machinery (ACM)},
}

@InProceedings{Luna2011,
  author    = {Luna, Francisco and Gonzalez-Alvarez, David L. and Chicano, Francisco and Vega-Rodriguez, Miguel A.},
  booktitle = {2011 11th International Conference on Intelligent Systems Design and Applications},
  title     = {On the scalability of multi-objective metaheuristics for the software scheduling problem},
  year      = {2011},
  month     = nov,
  pages     = {1110--1115},
  publisher = {IEEE},
  doi       = {10.1109/isda.2011.6121807},
}

@Article{Jozefowiez2008,
  author    = {Jozefowiez, Nicolas and Semet, Frédéric and Talbi, El-Ghazali},
  journal   = {European Journal of Operational Research},
  title     = {Multi-objective vehicle routing problems},
  year      = {2008},
  issn      = {0377-2217},
  month     = sep,
  number    = {2},
  pages     = {293--309},
  volume    = {189},
  doi       = {10.1016/j.ejor.2007.05.055},
  publisher = {Elsevier BV},
}

@Article{Dai2019,
  author    = {Dai, Penglin and Liu, Kai and Feng, Liang and Zhang, Haijun and Lee, Victor Chung Sing and Son, Sang Hyuk and Wu, Xiao},
  journal   = {IEEE Transactions on Intelligent Transportation Systems},
  title     = {Temporal Information Services in Large-Scale Vehicular Networks Through Evolutionary Multi-Objective Optimization},
  year      = {2019},
  issn      = {1558-0016},
  month     = jan,
  number    = {1},
  pages     = {218--231},
  volume    = {20},
  doi       = {10.1109/tits.2018.2803842},
  publisher = {Institute of Electrical and Electronics Engineers (IEEE)},
}

@Article{Tian2021,
  author    = {Tian, Ye and Su, Xiaochun and Su, Yansen and Zhang, Xingyi},
  journal   = {IEEE Transactions on Emerging Topics in Computational Intelligence},
  title     = {{EMODMI}: A Multi-Objective Optimization Based Method to Identify Disease Modules},
  year      = {2021},
  issn      = {2471-285X},
  month     = aug,
  number    = {4},
  pages     = {570--582},
  volume    = {5},
  doi       = {10.1109/tetci.2020.3014923},
  publisher = {Institute of Electrical and Electronics Engineers (IEEE)},
}

@InProceedings{Durillo2008,
  author    = {Durillo, Juan J. and Nebro, Antonio J. and Coello Coello, Carlos A. and Luna, Francisco and Alba, Enrique},
  booktitle = {2008 IEEE Congress on Evolutionary Computation (IEEE World Congress on Computational Intelligence)},
  title     = {A comparative study of the effect of parameter scalability in multi-objective metaheuristics},
  year      = {2008},
  month     = jun,
  pages     = {1893--1900},
  publisher = {IEEE},
  doi       = {10.1109/cec.2008.4631047},
}

@Article{Durillo2010,
  author    = {Durillo, J J and Nebro, A J and Coello, C A C and García-Nieto, José and Luna, F and Alba, E},
  journal   = {IEEE Transactions on Evolutionary Computation},
  title     = {A Study of Multiobjective Metaheuristics When Solving Parameter Scalable Problems},
  year      = {2010},
  issn      = {1089-778X},
  month     = aug,
  number    = {4},
  pages     = {618--635},
  volume    = {14},
  doi       = {10.1109/tevc.2009.2034647},
  publisher = {Institute of Electrical and Electronics Engineers (IEEE)},
}

@Article{Ishibuchi2015,
  author    = {Ishibuchi, Hisao and Akedo, Naoya and Nojima, Yusuke},
  journal   = {IEEE Transactions on Evolutionary Computation},
  title     = {Behavior of Multiobjective Evolutionary Algorithms on Many-Objective Knapsack Problems},
  year      = {2015},
  issn      = {1941-0026},
  month     = apr,
  number    = {2},
  pages     = {264--283},
  volume    = {19},
  doi       = {10.1109/tevc.2014.2315442},
  publisher = {Institute of Electrical and Electronics Engineers (IEEE)},
}

@Article{Aguirre2007,
  author    = {Aguirre, Hernán E. and Tanaka, Kiyoshi},
  journal   = {European Journal of Operational Research},
  title     = {Working principles, behavior, and performance of {MOEAs} on {MNK-landscapes}},
  year      = {2007},
  issn      = {0377-2217},
  month     = sep,
  number    = {3},
  pages     = {1670--1690},
  volume    = {181},
  doi       = {10.1016/j.ejor.2006.08.004},
  publisher = {Elsevier BV},
}

@Article{Deb2002,
  author    = {Deb, K. and Pratap, A. and Agarwal, S. and Meyarivan, T.},
  journal   = {IEEE Transactions on Evolutionary Computation},
  title     = {A fast and elitist multiobjective genetic algorithm: {NSGA-II}},
  year      = {2002},
  issn      = {1089-778X},
  month     = apr,
  number    = {2},
  pages     = {182--197},
  volume    = {6},
  doi       = {10.1109/4235.996017},
  publisher = {Institute of Electrical and Electronics Engineers (IEEE)},
}

@Article{Beume2007,
  author    = {Beume, Nicola and Naujoks, Boris and Emmerich, Michael},
  journal   = {European Journal of Operational Research},
  title     = {{SMS-EMOA}: Multiobjective selection based on dominated hypervolume},
  year      = {2007},
  issn      = {0377-2217},
  month     = sep,
  number    = {3},
  pages     = {1653--1669},
  volume    = {181},
  doi       = {10.1016/j.ejor.2006.08.008},
  publisher = {Elsevier BV},
}

@Article{Zhang2007,
  author    = {Qingfu Zhang and Hui Li},
  journal   = {IEEE Transactions on Evolutionary Computation},
  title     = {{MOEA/D}: A Multiobjective Evolutionary Algorithm Based on Decomposition},
  year      = {2007},
  issn      = {1089-778X},
  month     = dec,
  number    = {6},
  pages     = {712--731},
  volume    = {11},
  doi       = {10.1109/tevc.2007.892759},
  publisher = {Institute of Electrical and Electronics Engineers (IEEE)},
}

@Article{Stadler1979,
  author    = {Stadler, W.},
  journal   = {Journal of Optimization Theory and Applications},
  title     = {A survey of multicriteria optimization or the vector maximum problem, part {I}: 1776-1960},
  year      = {1979},
  issn      = {1573-2878},
  month     = sep,
  number    = {1},
  pages     = {1--52},
  volume    = {29},
  doi       = {10.1007/bf00932634},
  publisher = {Springer Science and Business Media LLC},
}

@Article{Laumanns2004,
  author    = {Laumanns, M. and Thiele, L. and Zitzler, E.},
  journal   = {IEEE Transactions on Evolutionary Computation},
  title     = {Running Time Analysis of Multiobjective Evolutionary Algorithms on Pseudo-Boolean Functions},
  year      = {2004},
  issn      = {1089-778X},
  month     = apr,
  number    = {2},
  pages     = {170--182},
  volume    = {8},
  doi       = {10.1109/tevc.2004.823470},
  publisher = {Institute of Electrical and Electronics Engineers (IEEE)},
}

\end{document}